%% file: main.tex
\pdfoutput=1

\documentclass[10pt,journal,compsoc]{IEEEtran}
\IEEEoverridecommandlockouts
\usepackage{cite}
\usepackage{amsmath,amssymb,amsfonts,bm}
\usepackage{algorithmic}
\usepackage{graphicx}
\usepackage{balance}
\usepackage{enumitem}
\usepackage{textcomp}
\usepackage{xcolor}
\usepackage{subfigure}
\usepackage{algorithm}
\usepackage{multirow}
\usepackage{url}
\usepackage{booktabs}
\usepackage{array}
\usepackage{caption}
\usepackage{colortbl}
\usepackage{balance}
\usepackage{float,color}
\usepackage{romannum}

\def\BibTeX{{\rm B\kern-.05em{\sc i\kern-.025em b}\kern-.08em
    T\kern-.1667em\lower.7ex\hbox{E}\kern-.125emX}}
\begin{document}
\def\method{UGNN}
\newcommand{\R}[1]{{\color{black}{\sc} #1}}
%
\title{Towards Semi-supervised Universal Graph Classification}
%
%
%
%

\author{Xiao Luo$^*$, Yusheng Zhao$^*$, Yifang Qin$^*$, Wei Ju$^\dagger$, and Ming Zhang$^\dagger$
\IEEEcompsocitemizethanks{
\IEEEcompsocthanksitem Xiao Luo is with Department of Computer Science, University of California, Los Angeles, USA. (e-mail: xiaoluo@cs.ucla.edu)
\IEEEcompsocthanksitem Yusheng Zhao, Wei Ju, Yifang Qin and Ming Zhang are with National Key Laboratory for Multimedia Information Processing, School of Computer Science, Peking University, Beijing, China. (e-mail: yusheng.zhao@stu.pku.edu.cn, juwei@pku.edu.cn, qinyifang@pku.edu.cn, mzhang\_cs@pku.edu.cn)
\IEEEcompsocthanksitem Xiao Luo, Yusheng Zhao and Yifang Qin contribute equally to this work.
\IEEEcompsocthanksitem Corresponding authors: Wei Ju and Ming Zhang.
\IEEEcompsocthanksitem The paper is partially supported by National Natural Science Foundation of China (NSFC Grant Number 62276002).
}
\thanks{Manuscript received April 19, 2005; revised August 26, 2015.}}

%
%

\markboth{Journal of \LaTeX\ Class Files,~Vol.~14, No.~8, August~2015}%
{Shell \MakeLowercase{\textit{et al.}}: Bare Advanced Demo of IEEEtran.cls for IEEE Computer Society Journals}
%



\IEEEtitleabstractindextext{%
\begin{abstract}
\input{1_abstract}

\end{abstract}

}

\maketitle
\input{2_introduction}
\input{3_related_work}
\input{4_method}

\input{5_experiment}

\input{6_conclusion}

\IEEEdisplaynontitleabstractindextext

%
\IEEEpeerreviewmaketitle

\ifCLASSOPTIONcompsoc
  \section*{Acknowledgments}
\else
  \section*{Acknowledgment}
\fi

The authors are grateful to the anonymous reviewers for critically reading this article and for giving important suggestions to improve this article.

\ifCLASSOPTIONcaptionsoff
  \newpage
\fi



%
\bibliographystyle{IEEEtran}
\bibliography{rec}

%

\begin{IEEEbiography}
[{\includegraphics[width=1in,height=1.25in]{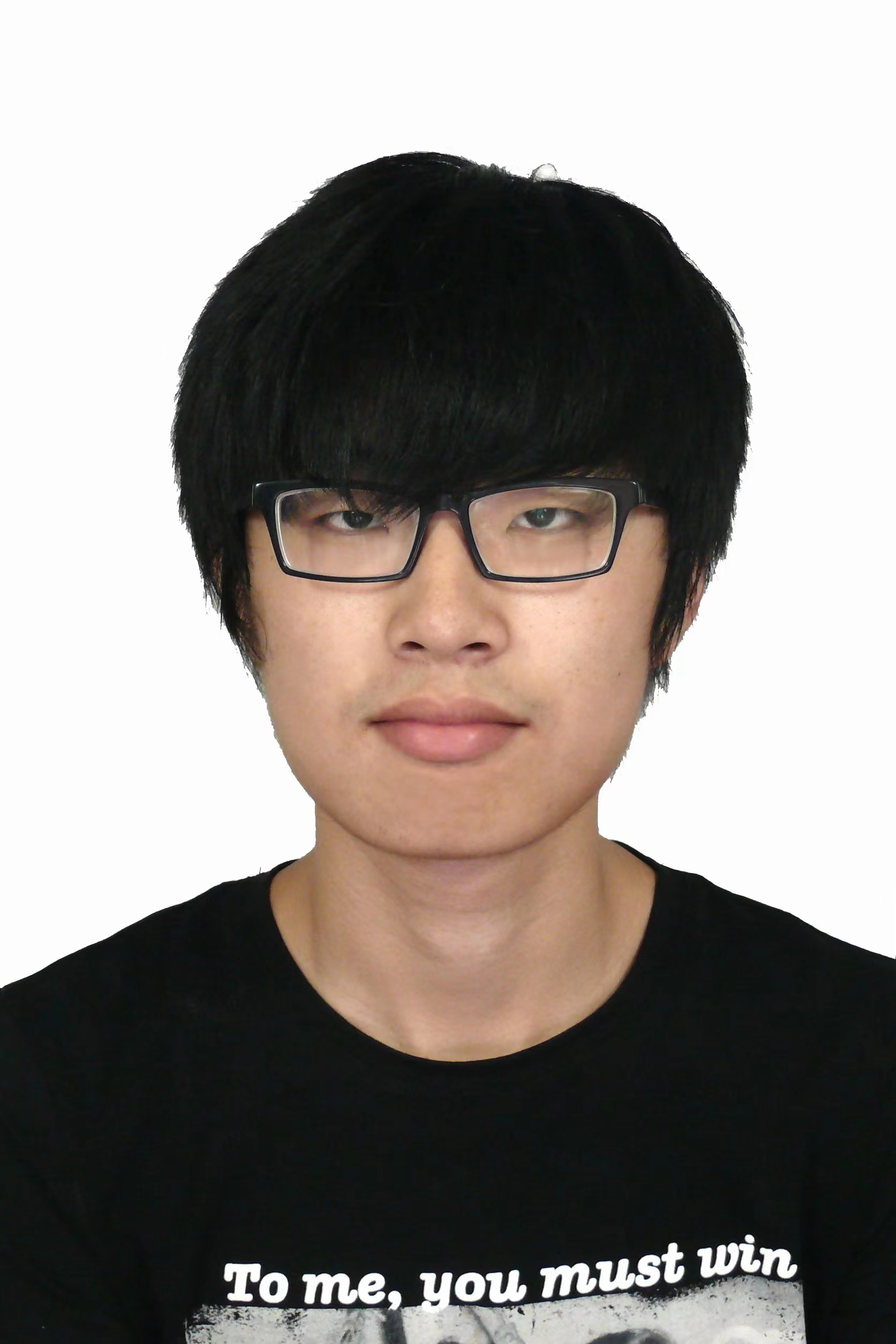}}]
{Xiao Luo} is a postdoctoral researcher in Department of Computer Science, University of California, Los Angeles, USA. Prior to that, he received the Ph.D. degree in School of Mathematical Sciences from Peking University, Beijing, China and the B.S. degree in Mathematics from Nanjing University, Nanjing, China, in 2017. 
His research interests includes machine learning on graphs, image retrieval, statistical models and bioinformatics. 
\end{IEEEbiography}

\begin{IEEEbiography}
[{\includegraphics[width=1in,height=1.25in]{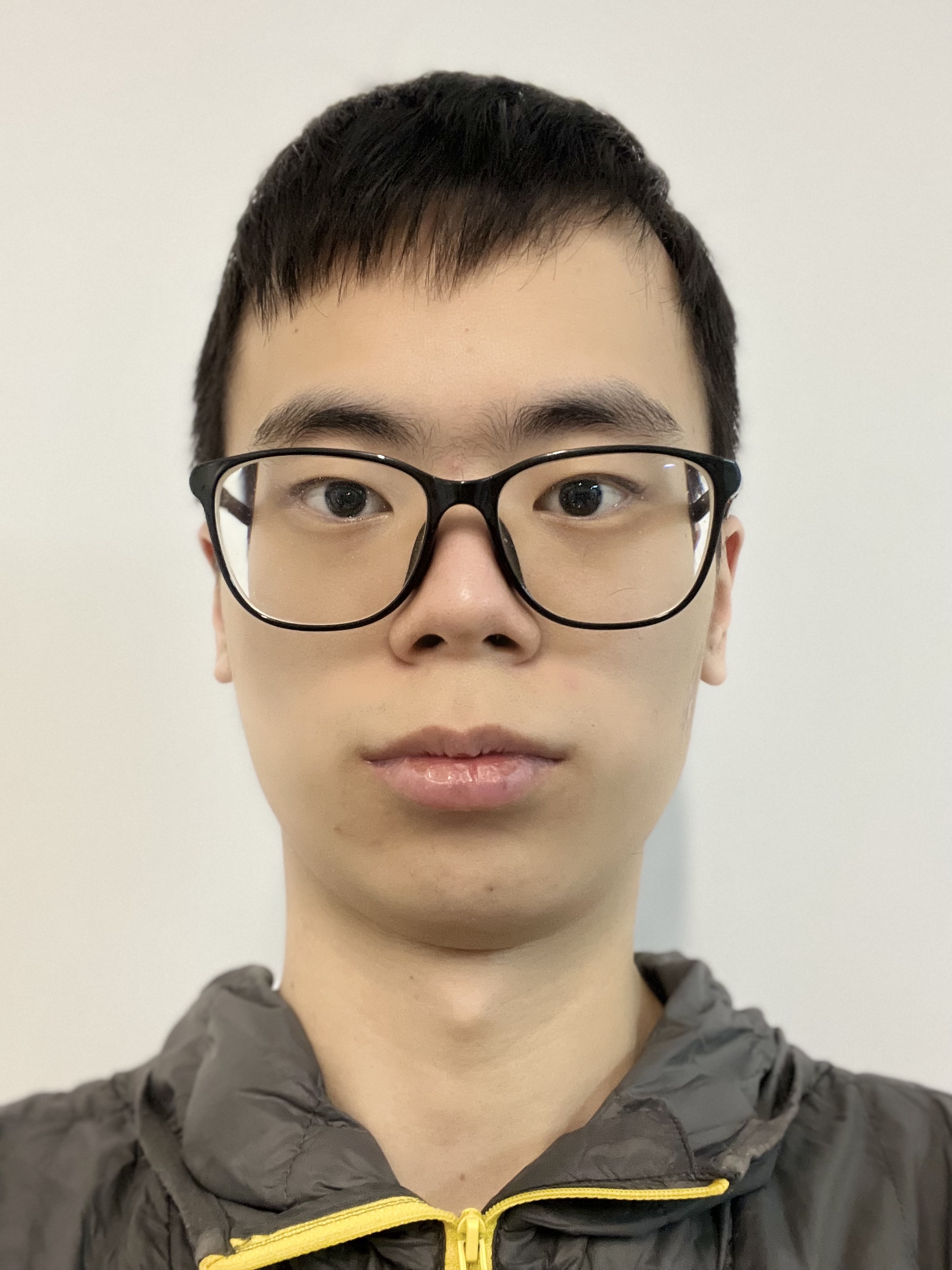}}]
{Yusheng Zhao} is a graduate student in School of Computer Science, Peking University, Beijing, China. His research interest includes machine learning with graphs, vision-and-language and adversarial learning.
\end{IEEEbiography}

\begin{IEEEbiography}
[{\includegraphics[width=1in,height=1.25in]{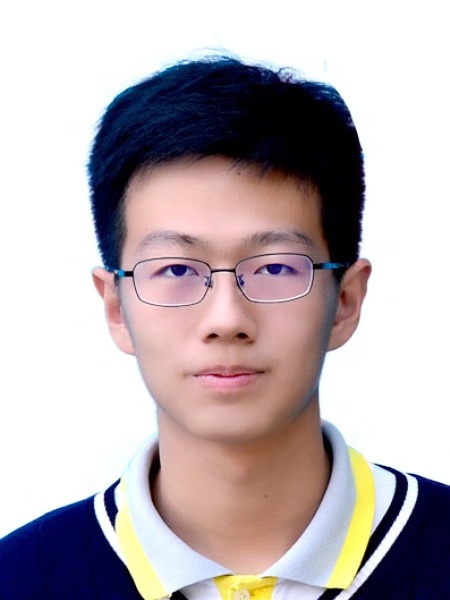}}]
{Yifang Qin} is an undergraduate student in School of EECS, Peking University, Beijing, China. His research interests include graph representation learning and recommender sysmtes.
\end{IEEEbiography}

\begin{IEEEbiography}
[{\includegraphics[width=1in,height=1.25in]{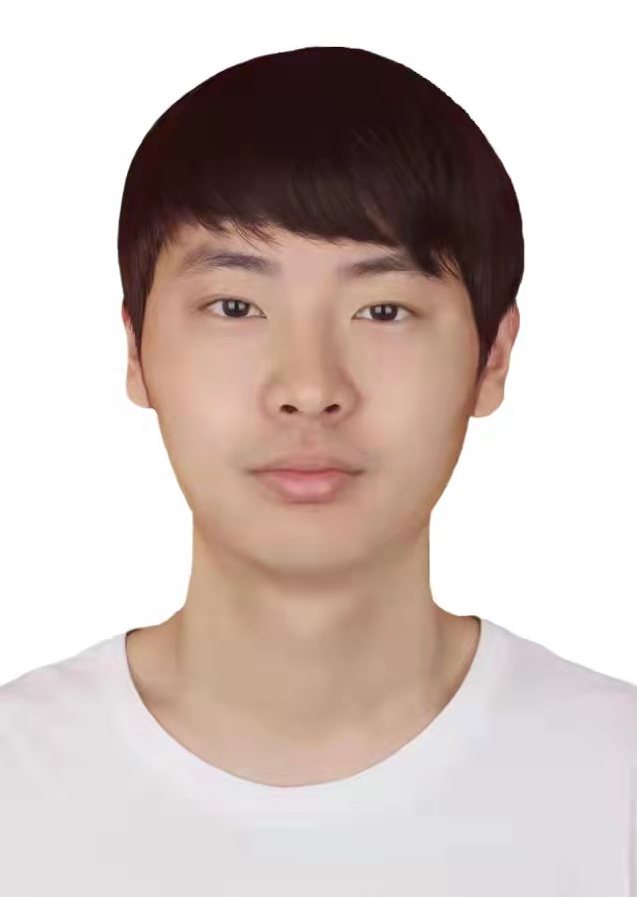}}]
{Wei Ju} is currently a postdoc research fellow in Computer Science at Peking University. Prior to that, he received his Ph.D. degree in Computer Science from Peking University, Beijing, China, in 2022. He received the B.S. degree in Mathematics from Sichuan University, Sichuan, China, in 2017. His current research interests lie primarily in the area of machine learning on graphs including graph representation learning and graph neural networks, and interdisciplinary applications such as drug discovery and recommender systems. He has published more than 20 papers in top-tier venues and has won the best paper finalist in IEEE ICDM 2022.
\end{IEEEbiography}

\begin{IEEEbiography}
[{\includegraphics[width=1in,height=1.25in]{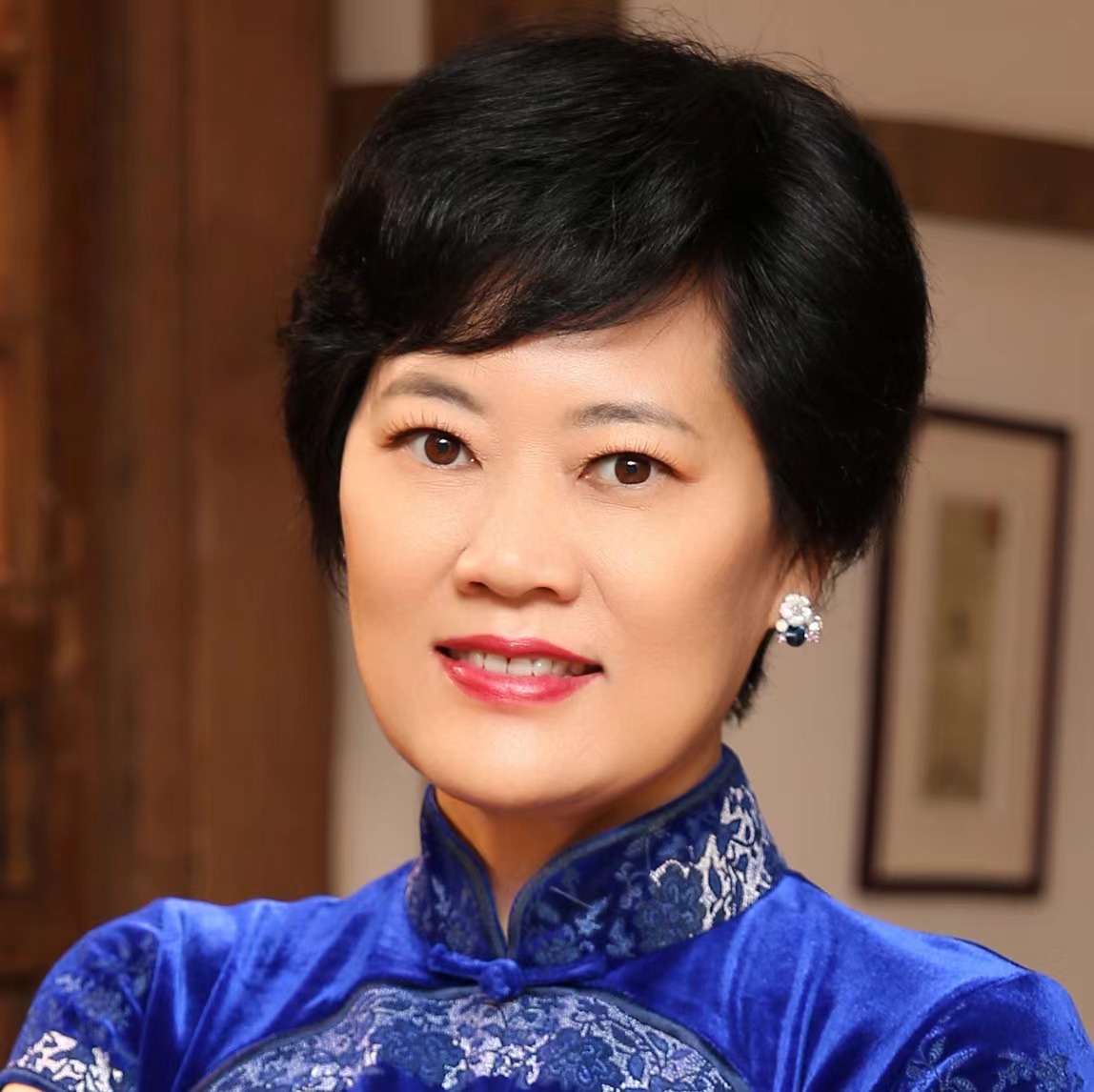}}]
{Ming Zhang} received her B.S., M.S. and Ph.D. degrees in Computer Science from Peking University respectively. She is a full professor at the School of Computer Science, Peking University. Prof. Zhang is a member of Advisory Committee of Ministry of Education in China and the Chair of ACM SIGCSE China. She is one of the fifteen members of ACM/IEEE CC2020 Steering Committee. She has published more than 200 research papers on Text Mining and Machine Learning in the top journals and conferences. She won the best paper of ICML 2014 and best paper nominee of WWW 2016. Prof. Zhang is the leading author of several textbooks on Data Structures and Algorithms in Chinese, and the corresponding course is awarded as the National Elaborate Course, National Boutique Resource Sharing Course, National Fine-designed Online Course, National First-Class Undergraduate Course by MOE China.
\end{IEEEbiography}

\end{document}

%% file: 1_abstract.tex
\R{Graph neural networks have pushed state-of-the-arts in graph classifications recently.} Typically, these methods are studied within the context of supervised end-to-end training, which necessities copious task-specific labels. However, in real-world circumstances, labeled data could be limited, and there could be a massive corpus of unlabeled data, even from unknown classes as a complementary. Towards this end, we study the problem of semi-supervised universal graph classification, which not only identifies graph samples which do not belong to known classes, but also classifies the remaining samples into their respective classes. This problem is challenging due to a severe lack of labels and potential class shifts. In this paper, we propose a novel graph neural network framework named \method{}, which makes the best of unlabeled data from the subgraph perspective.
To tackle class shifts, we estimate the certainty of unlabeled graphs using multiple subgraphs, which facilities the discovery of unlabeled data from unknown categories. Moreover, we construct semantic prototypes in the embedding space for both known and unknown categories and utilize posterior prototype assignments inferred from the Sinkhorn-Knopp algorithm to learn from abundant unlabeled graphs across different subgraph views. \R{Extensive experiments on six datasets verify the effectiveness of \method{} in different settings.}

%% file: 2_introduction.tex
\section{Introduction}

Graphs have garnered growing interest due to their capacity of portraying structured and relational data in a large range of domains. As one of the most prevalent graph machine learning problems, graph classification aims to predict the properties of whole graphs, which has widespread applications in visual and biological systems~\cite{hansen2015machine,ying2018graph,lee2019self,ying2018hierarchical}. In recent years, graph neural networks (GNNs) have exhibited promising performance in graph classification~\cite{lu2019molecular,schutt2017schnet,gilmer2017neural}, which usually follow the paradigm of message passing~\cite{kipf2017semi,zhong2021latent,shimin2021efficient,wang2021forecasting}. In detail, node representations are iteratively updated by aggregating neighborhood information, followed by a readout operation to generate graph representations. These graph representations can implicitly capture the structural topology in an end-to-end fashion, therefore facilitating downstream classification effectively.

\begin{figure}[t]
    \centering
    \includegraphics[width=0.48\textwidth,keepaspectratio=true]{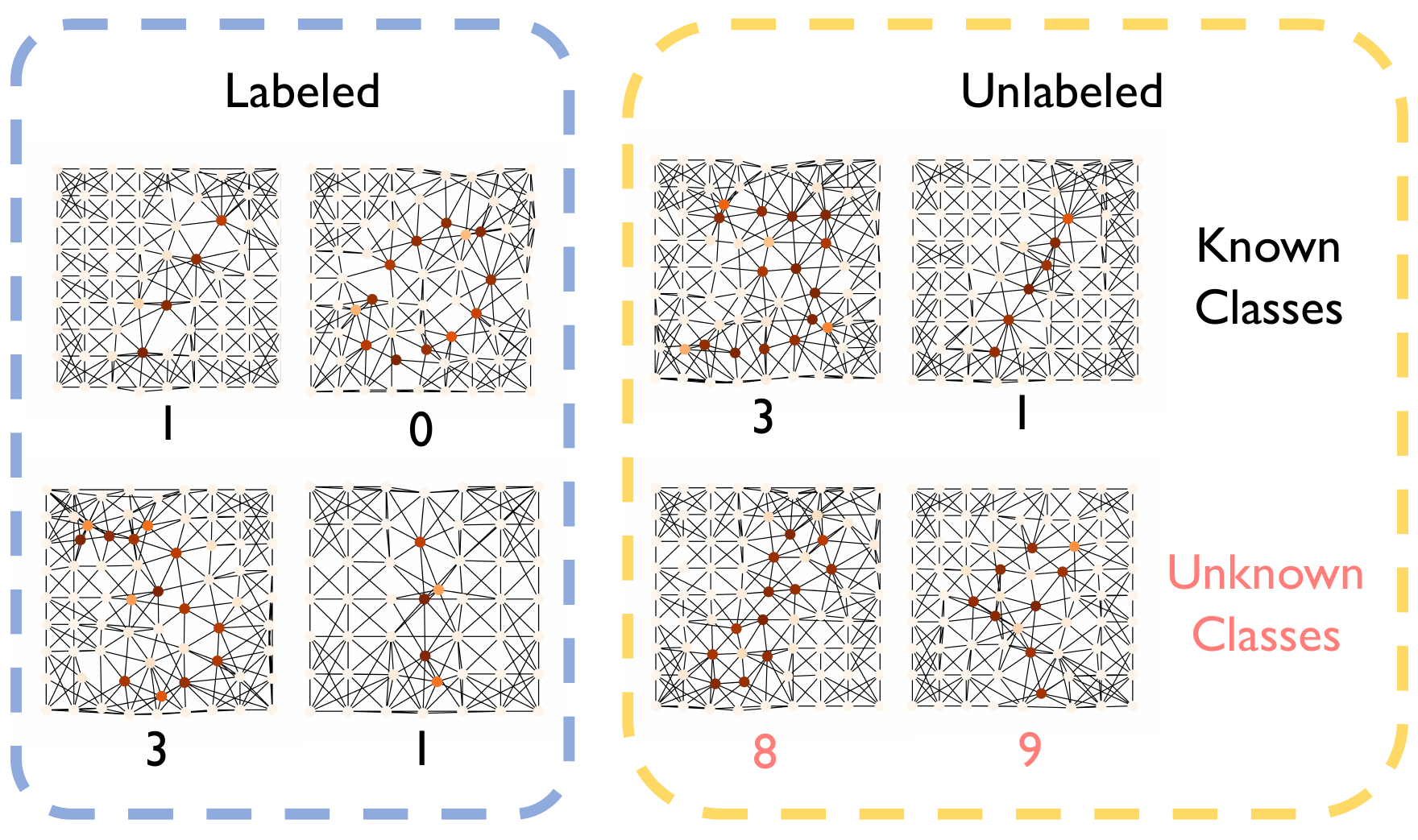}
    \caption{An illustration of our problem setting. We are given both labeled graphs and unlabeled graphs which could contain samples from unknown classes. }
    \label{fig:task}
\end{figure}

Despite their exceptional performance, GNNs are heavily reliant on copious task-specific labels while learning graph representation. In various real-world settings, however, large-scale data annotations could need a significant number of human resources~\cite{gilmer2017neural}. To address this, semi-supervised graph classification approaches have been proposed~\cite{li2019semi,sun2020infograph,hao2020asgn}, which use a huge corpus of unlabeled data to enhance the model performance in an efficient manner. These approaches presume that unlabeled graphs have the same distribution as labeled graphs, which would not be true in practice, particularly when labeled graphs make up a tiny portion of the whole dataset. For example, as in Fig. \ref{fig:task}, samples from classes '8' and '9' are unavailable in labeled data. To tackle this, in this research we investigate a more realistic problem named \textbf{semi-supervised universal graph classification}, where unlabeled data could belong to unknown classes. Here, two tasks must be carried out: (1) identifying graph samples that do not belong to known classes; and (2) categorizing the remaining samples into their respective classes. These out-of-distribution (OOD) graph samples (i.e., data from unknown classes) could be provided to experts, which increases the efficiency of data annotations.

In reality, this realistic graph classification would face the following essential difficulties: 
(1) \textbf{How can these OOD graph examples be detected in unlabeled data?}
The essence of this topic is to identify various samples that belong to unknown classes without adequate prior information. Typically, the bulk of conventional OOD contexts in computer vision assume that OOD samples are only engaged during assessment~\cite{hein2019relu,hendrycks2018deep,lee2018training}, but our problem focuses on their involvement during training. Even worse, this problem needs to deal with heterogeneous information in a large number of networks, i.e., node characteristics and structural topology, which makes discriminating between in-distribution (ID) and out-of-distribution (OOD) samples more challenging. (2) \textbf{How to overcome the label scarcity in the training data?} In reality, labeled data is scarce owing to the prohibitive expense of data annotation, but unlabeled data is abundant. Existing semi-supervised approaches often produce pseudo-labels to help optimize GNNs~\cite{sun2020infograph,hao2020asgn}. Nevertheless, these pseudo-labels could be biased and imbalanced, particularly when OOD graph samples exist. Note that resolving these obstacles might be mutually beneficial. On the one hand, the precise identification of OOD graph samples enables the best use of unlabeled ID graphs. On the other hand, exploring sufficient semantics from unlabeled data can assist in the discovery of OOD graph examples.

In this study, we present a novel framework \underline{U}niversal \underline{G}raph \underline{N}eural \underline{N}etwork (\method{}) that overcomes the aforementioned difficulties from the subgraph perspective. In particular, to produce graph-level representations, we first warmed up the message passing neural network with labeled graph samples. To combat class shift, we adopt a simple yet effective selection strategy, which samples numerous subgraphs to capture both prediction confidence and individual output uncertainty based on the calibration of GNNs. Our strategy computes both the average and the variance of prediction confidence scores among various subgraphs, and then sets an adaptive threshold to distinguish OOD samples from easy to hard. In addition, to make the most of unlabeled data, we construct graph prototypes in the embedding space for both known and unknown classes. Then, a semi-supervised prototype representation learning paradigm is developed, which utilizes the posterior prototype assignments from one subgraph view to supervise the semantics of unlabeled data from another view. The Sinkhorn-Knopp algorithm~\cite{caron2020unsupervised,asano2019self,zheng2021group} is involved to promise balanced and soft posterior distributions.
Our OOD sample selection technique and prototype-aware semi-supervised learning paradigm could mutually strengthen each other, enabling optimal use of unlabeled data. \R{To demonstrate the efficacy of our \method{}, we conduct extensive experiments on six benchmark graph classification datasets.} The results demonstrate that \method{} outperforms a number of state-of-the-art models in a range of settings.
The contributions of our work are summarized as follows:

\begin{itemize}
    \item \textbf{Problem Formalization:} We investigate the problem of semi-supervised universal graph classification, which facilities data annotation efficiency in real-world applications. 
    \item \textbf{Novel Methodologies:} We propose a simple yet effective method named \method{} to solve the problem. On the one hand, it captures individual output uncertainties by sampling multiple subgraphs to detect OOD graph samples. On the other hand, a semi-supervised prototype representation learning paradigm employs the posterior prototype assignments to supervise the semantics of unlabeled graphs across two subgraph views. 
    \item \textbf{Multifaceted Experiments:} \R{Extensive experiments on six graph classification datasets to validate the efficacy of the proposed \method{} in different settings.}
\end{itemize}

The related works are introduced in Section \ref{sec_related}. In Section \ref{sec_pre} and Section \ref{sec:method}, we describe the prior knowledge and the details of our \method{}, respectively. Section \ref{sec:experiment} offers extensive experimental results including quantitative comparisons, ablation studies, parameter sensitivity and visualization. In the end, we give a conclusion in Section \ref{sec:conclusion}.

%% file: 3_related_work.tex
\section{Related Work}\label{sec_related}
\subsection{Graph Neural Networks}

Graph neural networks (GNNs) have shown outstanding performance in relational data representation learning~\cite{zhang2020deep}, which has been extensively adopted in a number of applications, including node classification~\cite{guo2022learning,zhao2021graphsmote,liu2021relative}, link prediction~\cite{zhang2018link,cai2021line,zhu2021neural}, and anomaly detection~\cite{ma2021comprehensive,li2021cutpaste,deng2021graph}. Early efforts~\cite{defferrard2016convolutional,bruna2013spectral,henaff2015deep} usually utilize spectral GNNs based on the spectral graph theory, which begin with transferring graph signals into the embedding space, followed by spectral filters deduced from the graph Laplacian. Recent spatial methods have become the mainstream due to their lower computational complexity~\cite{kipf2017semi,velivckovic2018graph,xu2019powerful}. Typically, they adhere to the message passing paradigm, in which each node receives data from its neighbors, followed by an aggregation process that continually updates the node representations. GNNs have also been used regularly for graph classification. Typically, these methods use graph pooling functions to summarize node representations into graph-level representations~\cite{ying2018hierarchical,lee2019self}. For example, SAG Pooling~\cite{SAGpooling} utilizes the attention technique to preserve important nodes in a hierarchical fashion. Despite their promising performance, these methods are data-hungry whereas real-world applications often consist of limited labeled data and massive unlabeled data containing OOD graph samples. Towards this end, we investigate the semi-supervised universal graph classification problem, which not only identifies graph samples that do not belong to known classes but also classifies the remaining samples into their respective classes.

\subsection{Semi-supervised Graph Classification}

\R{Semi-supervised learning has received increasing attention in recent years. Pseudo-labeling is a popular technique which predicts the label distribution of unlabeled examples and selects confident samples for further guidance. 
For example, FlexMatch~\cite{zhang2021flexmatch} introduces class-specific adaptive thresholds to decide confident samples inspired by curriculum learning. Ada-CM~\cite{li2022towards} further utilizes contrastive learning to explore unconfident samples in the unlabeled set. Another line towards semi-supervised learning is consistency learning. FixMatch~\cite{sohn2020fixmatch} is a simple method which combines semi-supervised learning and consistency learning, achieving superior performance in this field. } 
The objective of semi-supervised graph classification is to predict graph properties using both labeled data and unlabeled data, which accounts for real-world label scarcity~\cite{li2019semi,sun2020infograph,hao2020asgn,you2020graph,ju2022kgnn}. Typically, previous approaches incorporate graph neural networks into semi-supervised learning techniques. Early attempts often use pseudo-labeling approaches~\cite{li2019semi}, which annotate unlabeled graphs using the classification model itself, and then add samples along with highly confident predictions to the training set. Unfortunately, these approaches may produce overconfident and skewed pseudo-labels, which could lead to an accumulation of errors during subsequent optimization. Considering the complexity of learning graph-level representations, recent approaches often use the multi-task learning framework where the teacher model attempts to learn discriminative graph representations, whereas the student model concentrates on the classification task~\cite{sun2020infograph,hao2020asgn}. However, these methods do not consider the realistic problem of potential OOD graph samples, which brings challenges in the sufficient exploration of unlabeled data. To tackle this, our \method{} employs a sample selection strategy, which captures prediction confidence as well as individual output uncertainty from the subgraph perspective.

\subsection{Out-of-distribution Detection}

Out-of-distribution (OOD) detection has been widely utilized in a variety of real-world applications~\cite{song2022learning}. Typically, this problem is addressed in two contexts, resulting in supervised methods and unsupervised methods. With access to identified OOD samples during optimization, supervised algorithms~\cite{hein2019relu,hendrycks2018deep,lee2018training} typically not only reduce the cross-entropy loss for ID samples, but also enforce the uniformity of the prediction distributions for OOD samples. In practical applications, it is difficult to locate OOD samples in advance. To circumvent this issue, unsupervised approaches do not use OOD samples during training~\cite{liang2017enhancing,sehwag2021ssd,vyas2018out}. They deploy post-hoc detectors based on distance measurements such as Mahalanobis distance~\cite{lee2018simple} after training classification models using ID samples. Despite the impressive performance of OOD detection in computer vision, its application to graphs remains underexplored. 
In contrast, we offer a novel graph neural network named \method{} that thoroughly both explores subgraphs to find OOD graph samples and learns from unlabeled graphs.

\begin{figure*}[t!]
    \centering
    \includegraphics[width=\textwidth]{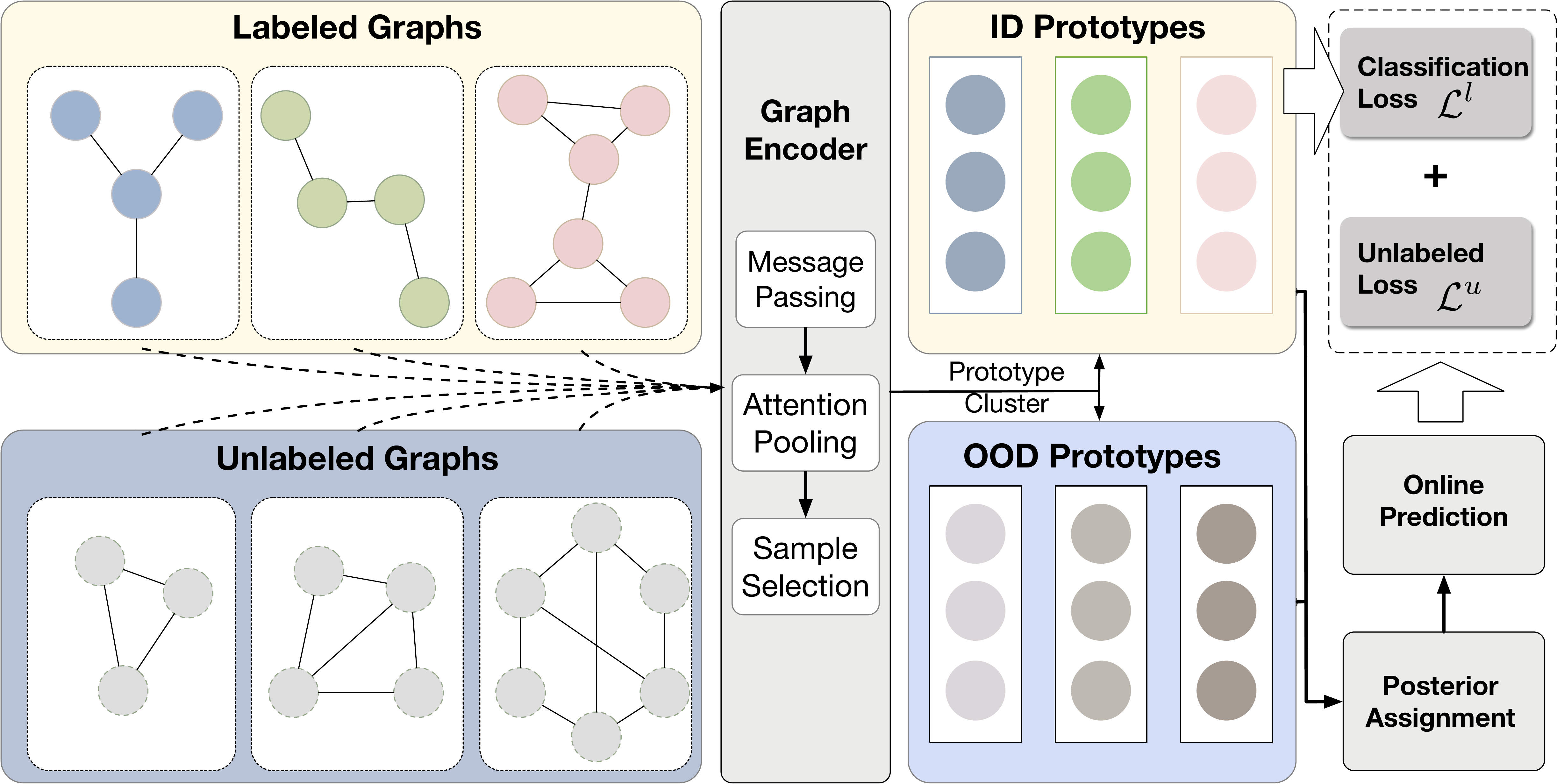}
    \caption{\R{Illustration of the proposed framework \method{}. Our \method{} is first warmed up using labeled data and then identifies OOD samples based on multiple subgraphs from each graph. Besides, \method{} constructs graph prototypes and compares the posterior prototype assignments with online prediction across different views.
    }}
    \label{fig:framework}
\end{figure*}

%% file: 4_method.tex
\section{Preliminary Knowledge}\label{sec_pre}
\subsection{Problem Formalization}
To begin with, we formally introduce the notations and the problem definition. Here a graph containing $n$ nodes is denoted as $G = (\boldsymbol{A}, \boldsymbol{X})$, where $\boldsymbol{A}\in \mathbb{R}^{n\times n}$ represents the adjacent matrix, $\boldsymbol{X} \in \mathbb{R}^{n\times d}$ represents the node attribute matrix and $d$ is the attribute dimension. In the problem of semi-supervised universal graph classification, we are given a training dataset $\mathcal{D}$ containing labeled graphs $\mathcal{D}^l=\{G^l_1,G^l_2,\cdots, G^l_{N^l} \}$ and unlabeled graphs $\mathcal{D}^u=\{G^u_1,G^u_2, \cdots, G^u_{N^u}\}$, where $G_{i}^l$ and $G_{j}^u$ represents the $i$-th labeled sample and the $j$-th unlabeled sample, respectively. The label set of the labeled data and the whole training data are denoted as $C^l$ and $C$, respectively. $y_i^l \in C^l$ denotes the label of $G_{i}^l$. Due to the potential label shifts, we have $C^l \subseteq C$. 

Our aim is to (1) identify graph samples which do
not belong to known classes, i.e., $\mathcal{S}=\{G_j^u| y_{j}^u\in C/C^l\}$ and (2) classify the remaining samples, i.e., $\mathcal{D}^u/\mathcal{S}$ into their corresponding classes in $C^l$.

\subsection{Message Passing Neural Networks}
We briefly introduce message passing neural networks, which are widely utilized to generate graph-level representations~\cite{kipf2017semi,velivckovic2018graph,xu2019powerful}. They usually utilize the neighborhood aggregation mechanism to explore topological information in an implicit fashion. The updating formulation at the $k$-th layer in a given graph $G$ is written as:
\begin{equation}\label{eq:message-passing}
\begin{aligned} \bm{v}_{N(v_i)}^{(k)} &=\operatorname{AGGREGATE}^{(k)}\left(\left\{\bm{v}_{i}^{(k-1)}: j \in \mathcal{N}(i)\right\}\right) \\ \bm{v}_{i}^{(k)} &=\operatorname{COMBINE}^{(k)}\left(\bm{v}_{i}^{(k-1)}, \bm{v}_{N(v_i)}^{(k)}\right) \end{aligned}
\end{equation}
where $\bm{v}_i^{(k)}$ is the representation of node $v_i$ at the $k$-th layer. $\operatorname{AGGREGATE}^{(k)}(\cdot)$ and $\operatorname{COMBINE}^{(k)}(\cdot)$ represent the aggregation and combination operators at the $k$-th layer, respectively. Finally, a global pooling operator is utilized to summarize all these node representations at the last layer, resulting in a graph-level representation:
\begin{equation}\label{eq:gp}
\bm{z}=\operatorname{GP}\left(\left\{\bm{v}_{i}^{(K)}\right\}_{i=1}^n\right),
\end{equation}
where $\operatorname{GP}(\cdot)$ represents a global pooling function. 

\section{Methodology}\label{sec:method}

This paper proposes a novel graph neural network framework named \method{} for semi-supervised universal graph classification. The core of our \method{} is to utilize subgraphs to sufficiently explore the semantics in unlabeled graphs. We first warm up our GNN-based encoder using labeled data to generate graph representations. To overcome label shift, we employ a sample selection strategy, which calculates confidence scores from the distribution viewpoint by sampling multiple subgraphs. To make the most of unlabeled data, we measure graph prototypes for both known and unknown classes, which 
can yield balanced and reliable prototype assignments by solving an optimization problem. Then, a semi-supervised graph prototype representation learning paradigm is presented, which utilizes the posterior prototype assignments from one subgraph view to supervise the semantics of unlabeled data from another view. More details can be illustrated in Fig. \ref{fig:framework}. 

\subsection{Graph Representation Learning}\label{sec:representation_learning}

Our model needs to extract topological information from both labeled and unlabeled graphs for the classification task with potential label shifts.
Therefore, learning effective graph representations is crucial for our problem. Towards this end, we leverage a message passing neural network to encode graphs into low-dimensional embeddings. In addition, a hierarchical graph pooling structure is adopted to explore local substructures in the graph.

In detail, given a graph sample $G=(\bm{A},\bm{X})$, we first utilize the message passing neural network in Equation \ref{eq:message-passing} to extract topological information, resulting in discriminative node representations, i.e., $\{\bm{v}_i^{(K)}\}$. Then, we follow the paradigm of TopK-based pooling by utilizing the attention mechanism to identify crucial nodes which will be kept. Here, we utilize a different encoder to produce an importance score vector $\bm{S}\in \mathbb{R}^{n\times 1}$ for nodes in the graph. The top $\lceil \rho n\rceil$ nodes will be kept by comparing the values in $\bm{S}$. Let $idx$ denote the index of kept nodes, and we derive pooled graph with the adjacent matrix $\tilde{\bm{A}}$ and the hidden embedding matrix $\tilde{\bm{H}}$ as follows:
\begin{equation}
\tilde{\bm{V}}=\bm{V}_{idx,:} \odot \bm{S}_{idx}, \tilde{\bm{A}}=\bm{A}_{idx, idx},
\end{equation}
where $\bm{V}_{idx,:}$ denotes stacked node representation matrix and $\odot$ represents the broadcasted Hadamard product~\cite{SAGpooling,ranjan2020asap}. $\bm{A}_{idx, idx}$ represents the row-wise and column-wise indexed matrix of $\bm{A}$. Then, the pooled graph is fed to the message passing neural network again, followed by the readout function as in Equation \ref{eq:gp}, producing a graph-level representation $\Phi(G)$. In this way, we summarize node semantics embedded in a hierarchical fashion, maximizing the model capacity for graph representation learning. 

Finally, we add an MLP classifier $\psi(\cdot)$ on top of the representations, which produces label distribution for each graph sample. Here the graph classification network is warmed up by merely labeled graph samples. Given a batch of labeled graphs $\mathcal{B}^l\subset \mathcal{D}^l$, the cross-entropy loss is employed for graph classification as follows:
\begin{equation}
\mathcal{L}^l=\frac{1}{\left|\mathcal{B}^{l}\right|} \sum_{G_{i}^l \in \mathcal{B}^{l}}-\log (\bm{y}^l_{i})^T  \psi(\Phi(G_i^l))
\end{equation}
where $\bm{y}^l_{i}$ denotes the one-hot vector of $y_i^l$. 

\subsection{Subgraph-based OOD detection}
The first aim of our problem is to detect OOD samples in unlabeled data. Intuitively, graph samples with shaped predicted label distributions are more likely to be ID samples. However, the serious label scarcity of labeled graph data could bring in biased and overconfident predictions. To tackle this, we incorporate multiple subgraphs sampled from the graph input into our OOD detection from the view of both confidence and model calibration.

In particular, we first introduce two perturbation strategies to generate subgraphs~\cite{you2020graph,you2021graph} as: (1) Edge deletion: we remove part of edges from a graph obeying an
i.i.d uniform distribution. (2) Node deletion: several nodes are removed at random, along with their connected edges. Then, a subgraph set
$S = \{\tilde{G}_1, \tilde{G}_2,\cdots, \tilde{G}_I\}$ can be generated for each graph $G$, where $I$ represents the number of subgraphs. We can derive the confidence score for each subgraph which is defined as the largest probability among the prediction distribution:
\begin{equation}
    s_r = \phi(f(\tilde{G}_r)),
\end{equation}
where $\phi:\mathbb{R}^{d} \rightarrow \mathbb{N}$ return the index of the maximum value. Here we measure the distribution of confidence score using a normal distribution, i.e., $N(\mu,\sigma^2)$.
The mean and the variance among the subgraph set are measured, i.e., $\hat{\mu}=\frac{1}{I}\sum_{i=1}^I s_i$ and $\hat{\sigma}^2=\frac{1}{I}\sum_{i=1}^I(s_i-\mu)^2$.
In this part, we adopt a hybrid strategy to determine the OOD samples based on both the mean and variance of the confidence distributions.

To begin with, samples with high confidence are more likely to belong to ID samples. Here we utilize the mean of the confidence scores to release the sample biases. \R{Intuitively, high variances for augmented views imply the prediction is not stable. In other words, their augmented views could be dissimilar to samples in the predicted class to generate a contradiction. Therefore, these samples could come from unseen classes, and we cannot identify them as ID samples. From a different view, compared with OOD samples, ID samples can be more resistant to the noise attack since they are correctly classified.} 
Therefore, we turn to model calibration.
In particular, the expected calibration error is directly related to the variance of the prediction empirically~\cite{rizve2021defense}. Therefore, we get more emphasis on the samples with low-variance confidence, which are resistant to the potential noise attack. Taking both factors into consideration, the final score for OOD detection is formulated as:
\begin{equation}
    \tilde{s} = \hat{\mu} - \hat{\sigma}.
\end{equation}
A higher score indicates a small probability of being OOD samples. Therefore, the set of ID samples is inferred as follows:
\begin{equation}\label{eq:generate_D}
    \mathcal{D}^{u,id} = \{G| \tilde{s}(G)>\delta \}
\end{equation}
where $\delta$ is a threshold to decide the portion of ID samples. Inspired by curriculum learning, we gradually raise the threshold to identify the OOD samples \R{from easy to hard}. For the $t$-th iteration, we have $\delta_t = \frac{t}{T}\delta$, where $T$ is the total number of iterations.

\subsection{Prototype-aware Semi-supervised Learning}
In addition, to overcome the label scarcity in the training data, we introduce a novel prototype-aware semi-supervised learning framework, which combines learning discriminative graph representations with semi-supervised learning.

\noindent\textbf{Prototype Initialization.} To begin with, we initialize prototypes for both ID samples and OOD samples. On the one hand, we take the average of graph representations from every known category in the embedding space. On the other hand, we cluster the graph representations of OOD samples selected before.

In detail, the prototype representation for the $c$-th category is defined as:
\begin{equation}\label{initalization}
    \bm{h}_c = \frac{\sum_{i=1}^{N^l} \bm{1}_{\bm{y}^l_{i}=c} \bm{z}_i}{\sum_{i=1}^{N^l} \bm{1}_{\bm{y}^l_{i}=c}},
\end{equation}
Then, the graph representations of OOD samples are clustered into $R$ parts, and the clustering center is denoted as $\bm{h}_{C+1},\cdots, \bm{h}_{C+R}$. These prototypes from ID and OOD samples jointly enhance semi-supervised learning for discriminative graph representations. 

\noindent\textbf{Semi-supervised Learning.} To learn from unlabeled data, we first generate their posterior prototype assignments to provide additional knowledge. To prevent generating overconfident distributions, prototype assignments are also inferred from the view of subgraphs, which would guide the semantic learning for unlabeled data. 

In detail, given a batch of unlabeled graphs $\{G_j^u\}_{j=1}^{B^u}$, we define matrix $\bm{Q}\in \mathbb{R}^{E\times B^u}=Eq\left(y=c \mid G_j^u\right)$ where $E=C+R$. We first stack the prototype representations and a batch of subgraph representations into the matrices $\bm{H}\in\mathbb{R}^{D\times E}$ and $\tilde{\bm{Z}}\in\mathbb{R}^{D\times B^u}$.
To obtain accurate and balanced posterior prototype assignments, we maximize the following objective as follows:
\begin{equation}\label{eq:obj}
\begin{aligned}
\mathcal{E}=&\left\langle\mathbf{H}^T\tilde{\bm{Z}}, \bm{Q}\right\rangle+\epsilon \mathcal{H}\left(\bm{Q}\right)+<\boldsymbol{f}, \bm{Q} \bm{1}_{B^u}-\frac{1}{E} \bm{1}_{E}>\\
&+\left\langle\boldsymbol{g}, \bm{Q}^{\top} \bm{1}_{E}-\frac{1}{B^u} \bm{1}_{B^u}\right\rangle
\end{aligned}
\end{equation}
where $<\mathbf{H}^T\tilde{\bm{Z}},\bm{Q}>$ returns the trace of $\tilde{\bm{Z}}^T\mathbf{H}\bm{Q}$ and $\mathcal{H}(\bm{Q})=-\sum_{i,j}\bm{Q}_{i,j}\log \bm{Q}_{i,j}$ denotes the entropy of the matrix. $\boldsymbol{f}\in \mathbf{R}^{B^u\times 1}$ and $\boldsymbol{g}\in \mathbf{R}^{E\times 1}$ are two adaptive Lagrange multipliers. In Equation \ref{eq:obj}, the first term maximizes the similarity between $\bm{Q}$ and $\mathbf{H}^T\tilde{\bm{Z}}$, which denotes the expected similarity scores between graph representations are their representations. The second term is a regularization term to maximize the entropy of $\bm{Q}$, which encourages the diversity of the posterior prototype assignment. $\epsilon$ is a temperature parameter to control the diversity. The last two terms are Lagrange constraints, which aim to produce $\bm{Q}$ with both row and column normalization to produce balanced posterior distributions~\cite{caron2020unsupervised,asano2019self}. 

To maximize Equation \ref{eq:obj}, we first calculate its gradient with respect to every element $\bm{Q}_{ij}$ as follows:
\begin{equation}
\frac{\partial \mathcal{E}}{\partial\left(\bm{Q}_{ij}\right)}=2[\mathbf{H}^T\tilde{\bm{Z}}]_{i j}-\epsilon \log \left(\bm{Q}\right)_{i j}+\boldsymbol{f}_{i}+\boldsymbol{g}_{j}
\end{equation}
Therefore, the closed solution is written as:
\begin{equation}
\mathbf{Q}^{*}=\operatorname{Diag}(\mathbf{u}) \exp \left(\frac{2\mathbf{H}^T\tilde{\bm{Z}}}{\epsilon}\right) \operatorname{Diag}(\mathbf{v})
\end{equation}
where $\mathbf{u}=\operatorname{diag}\left(\exp \left(\frac{\bm{f}}{\epsilon}\right)\right)$ and $\mathbf{v}=\operatorname{diag}\left(\exp \left(\frac{\bm{g}}{\epsilon}\right)\right)$. In practice, we recall the constraints embedded in Equation \ref{eq:obj}, i.e., $\bm{Q} \bm{1}_{B^u}-\frac{1}{E} \bm{1}_{E}=0$ and $\bm{Q}^{\top} \bm{1}_{E}-\frac{1}{B^u} \bm{1}_{B^u}=0$ and utilize the iterative Sinkhorn-Knopp algorithm~\cite{caron2020unsupervised,asano2019self,zheng2021group} to solve this, which repeatedly conducts row normalization and column normalization to $\exp \left(\frac{2\mathbf{H}^T\tilde{\bm{Z}}}{\epsilon}\right)$. Preliminary studies indicate that using three iterations would achieve incredible performance with less computational expense, and that soft target assignments have a higher performance than one-hot ones.

\begin{algorithm}[t]
\caption{Training Algorithm of \method{}}
\label{alg1}
\begin{algorithmic}[1]
\REQUIRE Labeled graphs $\mathcal{D}^l$; Unlabeled graphs $\mathcal{D}^u$;
\ENSURE Identify OOD graph samples $\mathcal{S}$ and generate the prediction for $\mathcal{D}^u/\mathcal{S}$;
\STATE Warm up the network using $\mathcal{D}^l$;
\FOR{$t=1,2, \cdots, T$}
\STATE Generate $\mathcal{D}^{u,id}$ using Equation \ref{eq:generate_D};
\STATE Initialize graph prototype representations using Equation \ref{initalization};
\REPEAT
\STATE Construct a mini-batch by sampling graphs from $\mathcal{D}^l$ and $\mathcal{D}^u$;
\STATE Generate the posterior distribution using the Sinkhorn-Knopp algorithm;
\STATE Calculate the final loss using Equation \ref{eq:final_loss};
\STATE Update the network parameters through back propagation;
\STATE Update the prototype representations using Equation \ref{eq:source-prototopy};
\UNTIL convergence
\ENDFOR

\end{algorithmic}
\end{algorithm}

{
\begin{table*}[h]
\centering
\caption{\R{Statistics of the datasets used in the experiments}.}
\label{tab:datasets}
\resizebox{\linewidth}{!}{
{
\begin{tabular}{c ccccccccc}
    \toprule
    \midrule  
    Dataset & \# Graphs & \# Classes & Avg. \# Nodes & Avg. \# Edges & \# Known & \# Unknown & Node feature & Low Ratio & High Ratio\\
    \midrule
    COIL-DEL & 3900 & 100 & 21.54 & 54.24 & 80 & 20 &  Coordinates & 50\% & 80\% \\\midrule
    Letter-High & 2250 & 15 & 4.67 & 4.50 & 10 & 5 &  Coordinates & 20\% & 40\%  \\\midrule
    MNIST & 55,000 & 10 & 70.6 & 564.5 & 7 & 3 & Pixel (Gray) + Coordinates  & 1\% & 3\% \\\midrule
    CIFAR10 & 45,000 & 10 & 117.6 & 941.2 & 7 & 3 & Pixel (RGB) + Coordinates & 30\% & 70\%  \\\midrule
    REDDIT-MULTI-12K & 11929 & 11 & 	391.41 & 456.89 & 7 & 4 & - & 30\% & 70\%  \\\midrule
    COLORS-3 & 	10500 & 11 & 	61.31 & 91.03 & 7 & 4 & Colors & 30\% & 80\%  \\
    \midrule
    \bottomrule
\end{tabular}
}
}
\end{table*}
}

Then, these prototype assignments are viewed as guidance to learn from unlabeled data. In particular, the cross-entropy loss objective is written as: 
\begin{equation}
\mathcal{L}^u=-\frac{1}{B^u}\sum_{j=1}^{B^u}\sum_{c=1}^{E} q\left(y=c \mid G_j^u\right) \log p\left(y=c \mid G^u_j\right)
\end{equation}
where online predictions are calculated by 
\begin{equation}\label{eq:contrast}
    p\left(y=c \mid G^u_j\right) = \frac{\exp{({{\tilde{\tilde{\bm{z}}}_j^u}^{\top}\bm{h}_c / \tau)}}}{\sum_{c'=1}^{E}\exp{({\tilde{\tilde{\bm{z}}}_j^u}^{\top}\bm{h}_c / \tau)}}
\end{equation}
Here $\tilde{\tilde{\bm{z}}}_j^u$ is the other subgraph representations of $G_j^u$ and $\tau$ is the temperature parameter. In this section, two different perturbations are involved for posterior prototype assignments and online predictions, which helps to capture the invariant semantics in unlabeled graphs with less bias. 

\noindent\textbf{Prototype Update.} Here, we update the prototype representations as the training process of semi-supervised learning. In particular, we aggregate all the graph representations which are close to each prototype using momentum update. 
Formally, we first generate the pseudo-labels for each unlabeled graph, i.e., $p_j=argmax_c \{\bm{h}_j^T\bm{z}_c \}$ and the updated prototypes are derived as:
\begin{equation}\label{eq:source-prototopy}
    \bm{h}_c \leftarrow \eta \bm{h}_c + (1-\eta) \frac{ \sum_{j=1}^{N^u}\bm{1}_{ p_j=c}\bm{z}_j^u}{\sum_{j=1}^{N^u}\bm{1}_{ p_j=c}}
\end{equation}
where $\eta$ is a momentum coefficient.

\subsection{Optimization}

In a nutshell, the overall training loss of our \method{} is summarized into:
\begin{equation}\label{eq:final_loss}
    \mathcal{L} = \mathcal{L}^l + \mathcal{L}^u
\end{equation}
We optimize the whole framework using mini-batch stochastic gradient descent. For every cycle, we update the set of ID unlabeled samples using curriculum learning and the number of total cycles is $T$. The detailed algorithm is summarized in Algorithm \ref{alg1}.

%% file: 5_experiment.tex
\section{Experiments}\label{sec:experiment}
In this section, exhaustive experiments are conducted on several datasets to demonstrate the effectiveness of the proposed \method{}. Particularly, we are interested in several research questions (RQs):
\begin{itemize}
    \item \textbf{{RQ~1}}: What is the overall performance of \method{} compared to baseline methods?
    \item \textbf{{RQ~2}}: What is the influence of Subgraph-based OOD Detection and Prototype-aware Semi-supervised Learning in the proposed task?
    \item \textbf{{RQ~3}}: Do the proposed model sensitive to hyperparameters like the number of clusters, the presumed number of OOD samples and the temperature in prototype-aware semi-supervised learning?
    \item \textbf{{RQ~4}}: Are there any visualization of classification results and learned representations to show the effectiveness of \method{}?
\end{itemize}

\subsection{Experimental Setup}
\subsubsection{Datasets}
In the experiments, we use six public graph datasets: COIL-DEL, Letter-high, MNIST, CIFAR10, REDDIT-MULTI-12K and COLORS-3. The detailed statistics of the datasets are listed in Table~\ref{tab:datasets}.

\textit{COIL-DEL}. COIL-DEL dataset~\cite{iam} is constructed by applying Harris corner detection and Delaunay Triangulation on images. The result of triangulation is converted to a graph, where nodes and edges represent ending points and lines.

\textit{Letter-high}. Letter-high dataset~\cite{iam} involves graph representations of 15 capital letters (\emph{i.e.} A, E, F, H, I, K, L, M, N, T, V, W, X, Y, Z). In the graph, nodes represent the endpoints of the drawing and (undirected) edges correspond to lines. The letters are highly distorted, which makes the task challenging.

\textit{MNIST}. MNIST dataset~\cite{dwivedi2020benchmarking} is constructed by extracting super-pixels (i.e., small regions of homogeneous intensity in every image) of the image. After the super-pixel extraction, a k-nearest-neighbor graph is then constructed to represent the image.

\textit{CIFAR10}. CIFAR10 dataset~\cite{dwivedi2020benchmarking} is constructed in the same way as MNIST. The difference between the two datasets is that CIFAR10 contains larger graphs with richer semantic meanings, which makes the classification task more challenging.

\R{\textit{REDDIT-MULTI-12K}. REDDIT-MULTI-12K dataset~\cite{henaff2015deep} contains graphs where nodes denote users and edges denote comments. The aim is to classify graphs into different communities.

\textit{COLORS-3}. COLORS-3 dataset~\cite{knyazev2019understanding} contains random graphs where each node shows one color from red, green and blue and we aim to capture the number of green nodes in each graph.}

\subsubsection{Evaluation Setting}
We split the classes into known classes and unknown classes, and the number of each is shown in Table~\ref{tab:datasets}. In the semi-supervised universal graph classification task, we only use \emph{some} of the labels in the known classes. All instances in the unknown classes are \emph{unlabeled}. In order to measure the models' performance in different application scenarios, \R{we adopt two settings with different labeling ratios (\emph{i.e.} High Ratio and Low Ratio) and high labeling ratio/low labeling ratio is between $1.6$ and $3$}. For example, on the MNIST dataset, we use 1\% of the labels and 3\% of the labels as high labeling ratio and low labeling ratio, respectively. \R{More details can be found in Table \ref{tab:datasets}.}

As for the evaluation metric, we use the classification accuracy as the default. In the proposed semi-supervised universal graph classification setting, a graph is classified correctly if and only if (\romannum{1}) the graph belongs to the source classes and the model predicts the correct label, or (\romannum{2}) the graph belongs to the novel classes and the model detects it as the novel instance.

{

\begin{table*}[t]
\centering
\caption{\R{Classification accuracy of different labeling ratios on six datasets. Our \method{} achieves the best performance significantly.} }
\label{tab:main_results}
\resizebox{\linewidth}{!}{
\R{
\begin{tabular}{c cc c cc c cc c cc c cc c cc}
    \toprule
    \midrule  
    \multirow{2}{*}{\textbf{Model}} & \multicolumn{2}{c}{\textbf{COIL-DEL}} && \multicolumn{2}{c}{\textbf{Letter-High}} && \multicolumn{2}{c}{\textbf{MNIST}} && \multicolumn{2}{c}{\textbf{CIFAR10}} && \multicolumn{2}{c}{\textbf{REDDIT-MULTI-12K}} && \multicolumn{2}{c}{\textbf{COLORS-3}}\\
    \cmidrule{2-3} \cmidrule{5-6} \cmidrule{8-9} \cmidrule{11-12} \cmidrule{14-15} \cmidrule{17-18}
    & Low Ratio & High Ratio && Low Ratio & High Ratio && Low Ratio & High Ratio && Low Ratio & High Ratio && Low Ratio & High Ratio && Low Ratio & High Ratio \\
    \midrule
    WL Kernel & 0.068$\pm$0.000 & 0.087$\pm$0.001 && 0.444$\pm$0.001 & 0.511$\pm$0.000 && 0.142$\pm$0.001 & 0.178$\pm$0.002 && 0.133$\pm$0.001 & 0.139$\pm$0.002 && 0.159$\pm$0.000 & 0.164$\pm$0.001 && 0.104$\pm$0.001 & 0.106$\pm$0.000 \\
    SP Kernel & 0.039$\pm$0.004 & 0.053$\pm$0.002 && 0.131$\pm$0.002 & 0.142$\pm$0.004 && 0.142$\pm$0.001 & 0.144$\pm$0.006 && 0.139$\pm$0.003 & 0.133$\pm$0.004 && 0.156$\pm$0.006 & 0.165$\pm$0.005 && 0.097$\pm$0.003 & 0.102$\pm$0.007 \\
    Graphlet Kernel & 0.035$\pm$0.002 & 0.049$\pm$0.001 && 0.129$\pm$0.002 & 0.156$\pm$0.004 && 0.118$\pm$0.001 & 0.139$\pm$0.001 && 0.101$\pm$0.000 & 0.110$\pm$0.000 && 0.114$\pm$0.002 & 0.129$\pm$0.003 && 0.090$\pm$0.001 & 0.095$\pm$0.004 \\
    \midrule
    GCN & 0.226$\pm$0.025 & 0.335$\pm$0.012 && 0.329$\pm$0.039 & 0.504$\pm$0.018 && 0.200$\pm$0.004 & 0.370$\pm$0.005 && 0.309$\pm$0.010 & 0.365$\pm$0.008 && 0.312$\pm$0.006 & 0.334$\pm$0.005 && 0.319$\pm$0.007 & 0.383$\pm$0.008 \\
    GraphSAGE & 0.377$\pm$0.032 & 0.397$\pm$0.015 && 0.409$\pm$0.007 & 0.582$\pm$0.004 && 0.196$\pm$0.002 & 0.420$\pm$0.019 && 0.329$\pm$0.006 & 0.351$\pm$0.002 && 0.244$\pm$0.004 & 0.262$\pm$0.004  && 0.313$\pm$0.016 & 0.340$\pm$0.019 \\
    GIN & 0.412$\pm$0.016 & 0.445$\pm$0.013 && 0.489$\pm$0.006 & 0.571$\pm$0.005 && 0.297$\pm$0.008 & 0.626$\pm$0.002 && 0.272$\pm$0.019 & 0.333$\pm$0.007 && 0.347$\pm$0.003 & 0.3583$\pm$0.004 && 0.311$\pm$0.006 & 0.335$\pm$0.008 \\
    \midrule
    ASAP & 0.553$\pm$0.006 & 0.601$\pm$0.023 && 0.567$\pm$0.017 & 0.609$\pm$0.006 && 0.482$\pm$0.006 & 0.541$\pm$0.004 && 0.386$\pm$0.009 & 0.395$\pm$0.004 && 0.335$\pm$0.008 & 0.367$\pm$0.012 && 0.321$\pm$0.003 & 0.343$\pm$0.009 \\
    SAG Pooling & 0.410$\pm$0.008 & 0.485$\pm$0.040 && 0.462$\pm$0.058 & 0.522$\pm$0.003 && 0.425$\pm$0.001 & 0.632$\pm$0.003 && 0.364$\pm$0.015 & 0.388$\pm$0.021 && 0.328$\pm$0.006 & 0.347$\pm$0.008 && 0.391$\pm$0.005 & 0.402$\pm$0.010 \\
    \midrule
    InfoGraph & 0.524$\pm$0.006 & 0.547$\pm$0.019 && 0.514$\pm$0.012 & 0.550$\pm$0.011 && 0.519$\pm$0.015 & 0.632$\pm$0.008 && 0.362$\pm$0.004 & 0.404$\pm$0.007 && 0.320$\pm$0.011 & 0.335$\pm$0.021 && 0.379$\pm$0.007 & 0.390$\pm$0.009 \\
    GraphCL & 0.563$\pm$0.013 & 0.606$\pm$0.024 && 0.562$\pm$0.006 & 0.635$\pm$0.005 && 0.498$\pm$0.015 & 0.700$\pm$0.013 && 0.373$\pm$0.006 & 0.410$\pm$0.008 && 0.327$\pm$0.004 & 0.360$\pm$0.005 && 0.382$\pm$0.007 & 0.399$\pm$0.004 \\
    GLA & 0.565$\pm$0.005 & 0.610$\pm$0.013 && 0.602$\pm$0.016 & 0.631$\pm$0.016 && 0.483$\pm$0.015 & 0.705$\pm$0.014 && 0.383$\pm$0.004 & 0.412$\pm$0.009 && 0.341$\pm$0.009 & 0.368$\pm$0.002 && 0.365$\pm$0.008 & 0.378$\pm$0.005 \\
    RGCL & 0.572$\pm$0.004 & 0.608$\pm$0.012 && 0.587$\pm$0.008 & \textbf{0.697}$\pm$0.012 && 0.475$\pm$0.013 & 0.714$\pm$0.011 && 0.356$\pm$0.009 & 0.406$\pm$0.005 && 0.345$\pm$0.012 & 0.366$\pm$0.007 && 0.390$\pm$0.008 & 0.409$\pm$0.004 \\
    \midrule
    \method{} (ours) & \textbf{0.594}$\pm$0.021 & \textbf{0.630}$\pm$0.007 && \textbf{0.640}$\pm$0.010 & 0.660$\pm$0.007 && \textbf{0.585}$\pm$0.004 & \textbf{0.730}$\pm$0.012 && \textbf{0.397}$\pm$0.004 & \textbf{0.420}$\pm$0.003 && \textbf{0.377}$\pm$0.006 & \textbf{0.383}$\pm$0.008 && \textbf{0.415}$\pm$0.003  & \textbf{0.434}$\pm$0.005 \\
    \midrule
    \bottomrule
\end{tabular}
}
}
\end{table*}
}
{
\begin{table*}[t]
\centering
\caption{\R{F1 score of OOD detection on six datasets. Our \method{} achieves the best performance significantly.} }
\label{tab:main_F1_AUC}

\resizebox{\linewidth}{!}{
\R{
\begin{tabular}{c cc c cc c cc c cc c cc c cc}
    \toprule
    \midrule  
    \multirow{2}{*}{\textbf{Model}} & \multicolumn{2}{c}{\textbf{COIL-DEL}} && \multicolumn{2}{c}{\textbf{Letter-High}} && \multicolumn{2}{c}{\textbf{MNIST}} && \multicolumn{2}{c}{\textbf{CIFAR10}} && \multicolumn{2}{c}{\textbf{REDDIT-MULTI-12K}} && \multicolumn{2}{c}{\textbf{COLORS-3}}\\
    \cmidrule{2-3} \cmidrule{5-6} \cmidrule{8-9} \cmidrule{11-12} \cmidrule{14-15} \cmidrule{17-18}
    & Low Ratio & High Ratio && Low Ratio & High Ratio && Low Ratio & High Ratio && Low Ratio & High Ratio && Low Ratio & High Ratio && Low Ratio & High Ratio \\
    \midrule
    GCN & 0.190$\pm$0.047 & 0.214$\pm$0.011 && 0.301$\pm$0.054 & 0.342$\pm$0.021 && 0.241$\pm$0.004 & 0.379$\pm$0.014&& 0.216$\pm$0.005 & 0.238$\pm$0.007 && 0.322$\pm$0.004 & 0.358$\pm$0.010 && 0.429$\pm$0.008 & 0.559$\pm$0.005 \\
    GraphSAGE & 0.207$\pm$0.052 & 0.216$\pm$0.054 && 0.389$\pm$0.011 & 0.342$\pm$0.032 && 0.185$\pm$0.012 & 0.194$\pm$0.006 && 0.273$\pm$0.012 & 0.289$\pm$0.009 && 0.300$\pm$0.002 & 0.322$\pm$0.006 && 0.497$\pm$0.017 & 0.620$\pm$0.013  \\
    GIN & 0.255$\pm$0.011 & 0.274$\pm$0.005 && 0.424$\pm$0.011 & 0.417$\pm$0.009 && 0.213$\pm$0.005 & 0.404$\pm$0.004 && 0.237$\pm$0.009 & 0.255$\pm$0.011 && 0.341$\pm$0.002 & 0.361$\pm$0.004 && 0.542$\pm$0.004 & 0.562$\pm$0.014 \\
    \midrule
    ASAP & 0.270$\pm$0.013 & 0.317$\pm$0.024 && 0.462$\pm$0.016 & 0.518$\pm$0.008 && 0.427$\pm$0.006 & 0.462$\pm$0.003 && 0.251$\pm$0.004 & 0.265$\pm$0.006 && 0.310$\pm$0.006 & 0.321$\pm$0.010 && 0.387$\pm$0.005 & 0.416$\pm$0.011 \\
    SAG Pooling & 0.222$\pm$0.016 & 0.250$\pm$0.012 && 0.516$\pm$0.027 & 0.404$\pm$0.008 && 0.401$\pm$0.002 & 0.412$\pm$0.002 && 0.245$\pm$0.012 & 0.286$\pm$0.024 && 0.352$\pm$0.012 & 0.364$\pm$0.015 && 0.597$\pm$0.009 & 0.633$\pm$0.005 \\
    \midrule
    InfoGraph & 0.232$\pm$0.014 & 0.250$\pm$0.016 && 0.459$\pm$0.006 & 0.511$\pm$0.007 && 0.443$\pm$0.019 & 0.506$\pm$0.015 && 0.264$\pm$0.004 & 0.265$\pm$0.006 && 0.317$\pm$0.012 & 0.365$\pm$0.015 && 0.547$\pm$0.016 & 0.562$\pm$0.019 \\
    GraphCL & 0.257$\pm$0.013 & 0.314$\pm$0.016 && 0.524$\pm$0.008 & 0.517$\pm$0.017 && 0.446$\pm$0.016 & 0.498$\pm$0.022 && 0.260$\pm$0.002 & 0.288$\pm$0.006 && 0.320$\pm$0.011 & 0.377$\pm$0.007 && 0.541$\pm$0.016 & 0.632$\pm$0.009 \\
    GLA & 0.281$\pm$0.012 & 0.296$\pm$0.041 && 0.520$\pm$0.029 & 0.526$\pm$0.005 && 0.440$\pm$0.008 & 0.509$\pm$0.073 && 0.291$\pm$0.009 & 0.294$\pm$0.004 && 0.356$\pm$0.003 & 0.383$\pm$0.004 && 0.570$\pm$0.009 & 0.577$\pm$0.012 \\
    RGCL & 0.282$\pm$0.003 & 0.305$\pm$0.018 && 0.502$\pm$0.033 & 0.520$\pm$0.015 && 0.439$\pm$0.011 & 0.510$\pm$0.007 && 0.285$\pm$0.014 & 0.237$\pm$0.009 && 0.326$\pm$0.008 & 0.365$\pm$0.010 && 0.562$\pm$0.013 & 0.603$\pm$0.004 \\
    \midrule
    \method{} (ours) & \textbf{0.319}$\pm$0.088 & \textbf{0.352}$\pm$0.025 && \textbf{0.567}$\pm$0.015 & \textbf{0.563}$\pm$0.004 && \textbf{0.458}$\pm$0.005 & \textbf{0.535}$\pm$0.006 && \textbf{0.313}$\pm$0.006 & \textbf{0.297}$\pm$0.007 && \textbf{0.428}$\pm$0.014 & \textbf{0.443}$\pm$0.008 && \textbf{0.608}$\pm$0.007 & \textbf{0.654}$\pm$0.004 \\
    \midrule
    \bottomrule
\end{tabular}
}
}
\end{table*}
}

\subsubsection{Baseline Methods}
\R{We compare the proposed \method{} with a wealth of baselines, ranging from kernel-based approaches to graph neural networks and semi-supervised graph classification methods.} The detailed baseline methods are described as follows:

\textit{Graph Kernel Methods.}
In the experiments, we take three graph kernels for comparison, including:
\begin{itemize}
    \item Weisfeiler-Lehman (WL) Kernel~\cite{wl-kernel} that utilizes the Weisfeiler-Lehman algorithm to generate features of nodes that are compared across graphs.
    \item Shortest-Path (SP) Kernel~\cite{shortest-path-kernel} that decomposes graph samples into shortest paths and contrasts pairs of the shortest paths based on the lengths.
    \item Graphlet Kernel~\cite{graphlet-kernel} that counts the graphlets in the input graphs and generates features according to their occurrence.
\end{itemize}
We use labeled data to fit a Support Vector Machine (SVM) with graph kernels and then make predictions with this SVM classifier.

\textit{Graph Convolutional Neural Networks.} We also adopt a variety of graph convolutional layers, including Graph Convolutional Network (GCN)~\cite{GCN}, GraphSAGE~\cite{GraphSAGE}, Graph Isomorphic Network (GIN)~\cite{GIN}. When applying different graph convolutions, we use TopK Pooling~\cite{topk-pooling} as the default graph pooling method.

\textit{Graph Pooling Methods.} As for graph pooling methods, we select three graph pooling methods as baselines. Specifically, we use:
\begin{itemize}
    \item TopK Pooling~\cite{topk-pooling} that is described in Sec~\ref{sec:representation_learning}.
    \item Self-Attention Graph Pooling (SAG Pooling)~\cite{SAGpooling} that is on basis of self-attention to jointly consider both node features and graph topology.
    \item Adaptive Structure Aware Pooling (ASAP)~\cite{ASAP} that utilizes self-attention and learns soft cluster assignments for each node to pool the graph.
\end{itemize}
When comparing different graph pooling methods, we keep graph convolution method as the default GIN.

\textit{Semi-supervised Graph Classification Methods.} Contrastive learning is widely used in graph classification in semi-supervised settings. Therefore, we select three graph contrastive learning methods and one knowledge distillation method as our baselines.
\begin{itemize}
    \item \R{InfoGraph~\cite{sun2019infograph} that incorporates a teacher encoder and a student encoder trained in supervised and self-supervised manners, respectively. Their knowledge is transferred by maximizing the mutual information.}
    \item GraphCL~\cite{GraphCL} that adopts graph augmentations and normalized temperature-scaled cross entropy loss (NT-xent) to learn generalizable, transferable and robust representations using contrastive learning.
    \item GLA~\cite{GLA} that designs label-invariant augmentations in the representation space, and achieves promising results in semi-supervised graph classification task.
    \R{\item RGCL~\cite{li2022let} that utilizes invariant rationale discovery to separate the graph into two parts. These two part would be fed into the contrastive learning framework to learn effective graph representations.}
\end{itemize}

\subsubsection{Implementation Details}
For graph representation learning, we use GIN convolution~\cite{GIN} as default. In subgraph-based OOD detection, we obtain the subgraphs by randomly deleting 20\% of the nodes and their corresponding edges. We extract a total of 3 subgraphs from the original graph (\emph{i.e.} $I=3$) to empirically compute the mean and variance. For prototype-aware semi-supervised learning, the OOD samples are clustered into 3 parts (\emph{i.e.} $R=3$). We set $\epsilon$ in Eq.~\ref{eq:obj} to 0.05 and the softmax temperature $\tau$ in Eq.~\ref{eq:contrast} to 0.1 following~\cite{caron2020unsupervised}. The momentum coefficient $\eta$ is set to $0.99$ as in~\cite{he2020momentum}. During optimization, we also use an additional supervised contrastive loss function in~\cite{khosla2020supervised} to help the training. The proposed \method{} is implemented with PyTorch and can be trained on an NVIDIA RTX GPU. We train the model for 100 epochs in which the first 50 epochs are used to warm-up the model with labeled data. We use Adam optimizer~\cite{kingma2014adam} with a learning rate of 0.001 and the batch size is set to 256.

\subsection{Main Results (RQ~1)}
\R{The classification accuracy and F1 score of \method{} in comparison with various baseline methods are shown in Table~\ref{tab:main_results} and Table~\ref{tab:main_F1_AUC}. }From the results, we have several observations.

Firstly, the proposed \method{} achieves a consistent improvement compared to all baseline methods in both low labeling ratio (Low Ratio columns in the table) and high labeling ratio (High Ratio columns in the table) scenarios on all four datasets. The significant improvement shows that the effectiveness of the proposed Subgraph-based OOD Detection and Prototype-aware Semi-supervised Learning. More specifically, we attribute the performance gain to the following aspects: (\romannum{1}) \method{} is better at detecting OOD samples. While baseline methods use prediction confidence as the metric for classifying known and unknown categories, our model tackles this problem from a subgraph perspective, which yields more robust OOD detection. (\romannum{2}) The proposed method provides consistent classification benefiting from learning both known and unknown prototypes, whereas baseline methods focus mainly on known categories. (\romannum{3}) The two components benefit each other. Better out-of-distribution detection provides more accurate information for semi-supervised learning with prototypes. Conversely, more consistent and robust representations helps with finding OOD samples.

Secondly, the proposed \method{} experiences more significant improvements in the low labeling ratio cases (\emph{e.g.} \textbf{8.96}\% absolute improvement on MNIST with low labeling ratio, compared to \textbf{2.59}\% absolute improvement with high labeling ratio). This shows that our \method{} is more helpful in the face of label scarcity, which is more common in real-world scenarios.

Thirdly, traditional graph kernel based methods~\cite{wl-kernel, shortest-path-kernel, graphlet-kernel} generally perform worse than graph neural network methods. One exception can be found in the Letter-high dataset, in which Weisfeiler-Lehman kernel achieves comparable accuracy with GNN-based methods. A possible explanation is that the Letter-high dataset contains smaller graphs (4.67 nodes in a graph on average) and relatively simple structures. 
Graph Neural Networks (GNNs)~\cite{GCN, GraphSAGE, GIN, ASAP, topk-pooling, SAGpooling} boost the performance in graph classification via learning deep representations of graphs as well as utilizing node attributes. Despite their performance gain, they lack the ability to utilize unlabeled data and fall short in detecting OOD samples.
\R{Semi-supervised graph classification methods~\cite{sun2020infograph, GraphCL, GLA, li2022let} make use of unlabeled data and thus improve the accuracy. However, they are still weak when dealing with graphs in unknown classes.}
In contrast, our model detects OOD samples based on subgraphs and learns the prototypes for both known and unknown classes in the semi-supervised setting, which further boosts the performance.

Finally, the overall performance, as well as the model's improvement, of MNIST is higher than CIFAR10, given that we use less data in MNIST than in CIFAR10. The reason is two-fold. Firstly, even for their counterparts in the computer vision domain, CIFAR10 is harder than MNIST. More importantly, the two graph datasets are constructed using super-pixels, which inevitably causes information loss with regard to the detail in the original image. Since the recognition of hand-written digits relies more on the global shape (which is less affected), it is reasonable that graph neural networks perform better on the MNIST dataset.

{
\begin{table*}[t]
\centering
\caption{\R{Ablation studies on all the datasets. Our full model achieves the best performance consistently.}}
\label{tab:ablation}
\resizebox{\linewidth}{!}{
\R{
\begin{tabular}{c c cc c cc c cc c cc c cc c cc}
    \toprule
    \midrule  
    \multirow{2}{*}{\textbf{Experiment}} && \multicolumn{2}{c}{\textbf{MNIST-low}} && \multicolumn{2}{c}{\textbf{MNIST-high}} && \multicolumn{2}{c}{\textbf{COIL-DEL-low}} && \multicolumn{2}{c}{\textbf{COIL-DEL-high}} && \multicolumn{2}{c}{\textbf{CIFAR10-low}} && \multicolumn{2}{c}{\textbf{CIFAR10-high}}\\
    \cmidrule{3-4} \cmidrule{6-7} \cmidrule{9-10} \cmidrule{12-13} \cmidrule{15-16} \cmidrule{18-19} 
    && Overall & Unknown && Overall & Unknown && Overall & Unknown && Overall & Unknown && Overall & Unknown && Overall & Unknown\\
    \midrule
    \method{} && \textbf{0.585} & \textbf{0.485} && \textbf{0.730} & \textbf{0.585} && \textbf{0.594} & \textbf{0.313} && \textbf{0.630} & \textbf{0.337} && \textbf{0.397} & \textbf{0.278} && \textbf{0.420} & \textbf{0.290} \\
    \midrule
    w/o S && 0.525 & 0.381 && 0.646 & 0.500 && 0.507 & 0.216 && 0.572 & 0.356 && 0.383 & 0.261 && 0.377 & 0.293 \\
    \midrule
    w/o P && 0.488 & 0.407 && 0.694 & 0.554 && 0.424 & 0.252 && 0.503 & 0.255 && 0.367 & 0.266 && 0.353 & 0.286 \\
    \midrule
    w/o KP && 0.490 & 0.472 && 0.723 & 0.561 && 0.467 & 0.268 && 0.525 & 0.313 && 0.349 & 0.247 && 0.359 & 0.251\\
    \midrule
    w/o UP && 0.530 & 0.455 && 0.713 & 0.567 && 0.475 & 0.262 && 0.549 & 0.342 && 0.340 & 0.238 && 0.371 & 0.247 \\
    \midrule
    \bottomrule
\end{tabular}
}
}

\end{table*}

\begin{figure*}[ht]
    \centering    \includegraphics[width=0.9\textwidth,keepaspectratio=true]{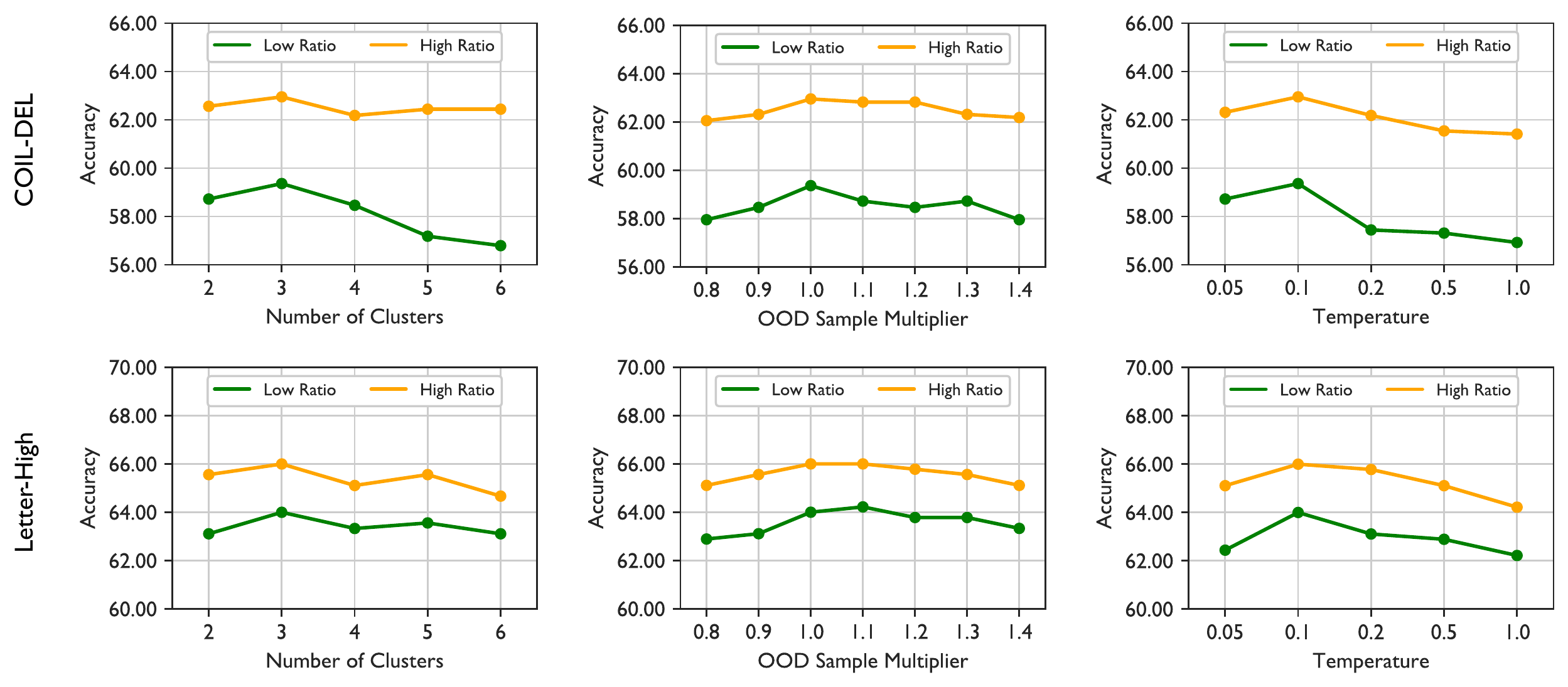}
    \caption{The parameter sensitivity experiments of \method{} in both low labeling ratio and high labeling ratio cases on two datasets (\emph{i.e.} COIL-DEL and Letter-high). The first column studies the number of novel clusters in Prototype-aware Semi-supervised Learning (\emph{i.e.} the number of prototypes of unknown classes $R$). The middle column shows the influence of the presumed number of OOD samples. The last column focuses on the temperature in Eq.~\ref{eq:contrast}.}\label{fig:sensitivity}
\end{figure*}
\subsection{Ablation Studies (RQ~2)}
In this subsection, we perform ablation studies to verify the effectiveness of Subgraph-based OOD Detection and Prototype-aware Semi-supervised Learning in \method{}. The ablated results are listed in Table~\ref{tab:ablation}, where \R{we compare the prediction accuracy of our model and several variants in both low labeling ratio (Low Ratio) and high labeling ratio (High Ratio) scenarios on all the datasets}. Moreover, in order to understand the effects of these components in the face of OOD samples, we also report the classification accuracy of unknown classes. Specifically, we compare \method{} to the model (\romannum{1}) without Subgraph-based OOD Detection (w/o S), (\romannum{2}) without Prototype-aware Semi-supervised Learning (w/o P), (\romannum{3}) without prototypes of known classes (w/o KP), and (\romannum{4}) without prototypes of unknown classes (w/o UP).

As can be seen from the results, removing each component results in performance drop in both low labeling ratio and high labeling ratio cases for both overall performance and OOD detection. This demonstrates the effectiveness of Subgraph-based OOD Detection and Prototype-aware Semi-supervised Learning. Concretely, we have several observations listed as follows:
\begin{itemize}
    \item Without effective OOD detection, the model fails to achieve satisfactory accuracy. \R{As we can see from the second line of Table~\ref{tab:ablation}, removing subgraph-based OOD detection causes catastrophic deterioration in classification accuracy of unknown classes (\emph{i.e.} 48.47\%$\to$38.13\% and 58.53\%$\to$49.97\% on MNIST)}.
    \item Without prototype-aware semi-supervised learning (w/o P, third line), the prediction accuracy falls. \R{Furthermore, it is worth noting that this component is more important when the label is scarce, which is demonstrated by a more significant decline in the low labeling ratio case (\emph{e.g.} 9.68\% overall accuracy decline in the low ratio case compared to 3.64\% in high ratio case on MNIST)}. This result is reasonable in that semi-supervised learning aims to utilize unlabeled data, which plays a more important role when labels are scarce.
    \item Removing either prototypes of known classes (w/o KP, fourth line) or prototypes of unknown classes (w/o UP, fifth line) also hinders the performance, but to a less extent compared to removing all the components (w/o P, third line). This is especially true when the model is provided with more labels. The mitigated performance drop suggests that the prototypes under the semi-supervised learning framework have enough representation power to partially supplant each other.
\end{itemize}

\begin{figure*}[ht]
    \centering
    \includegraphics[width=\textwidth,keepaspectratio=true]{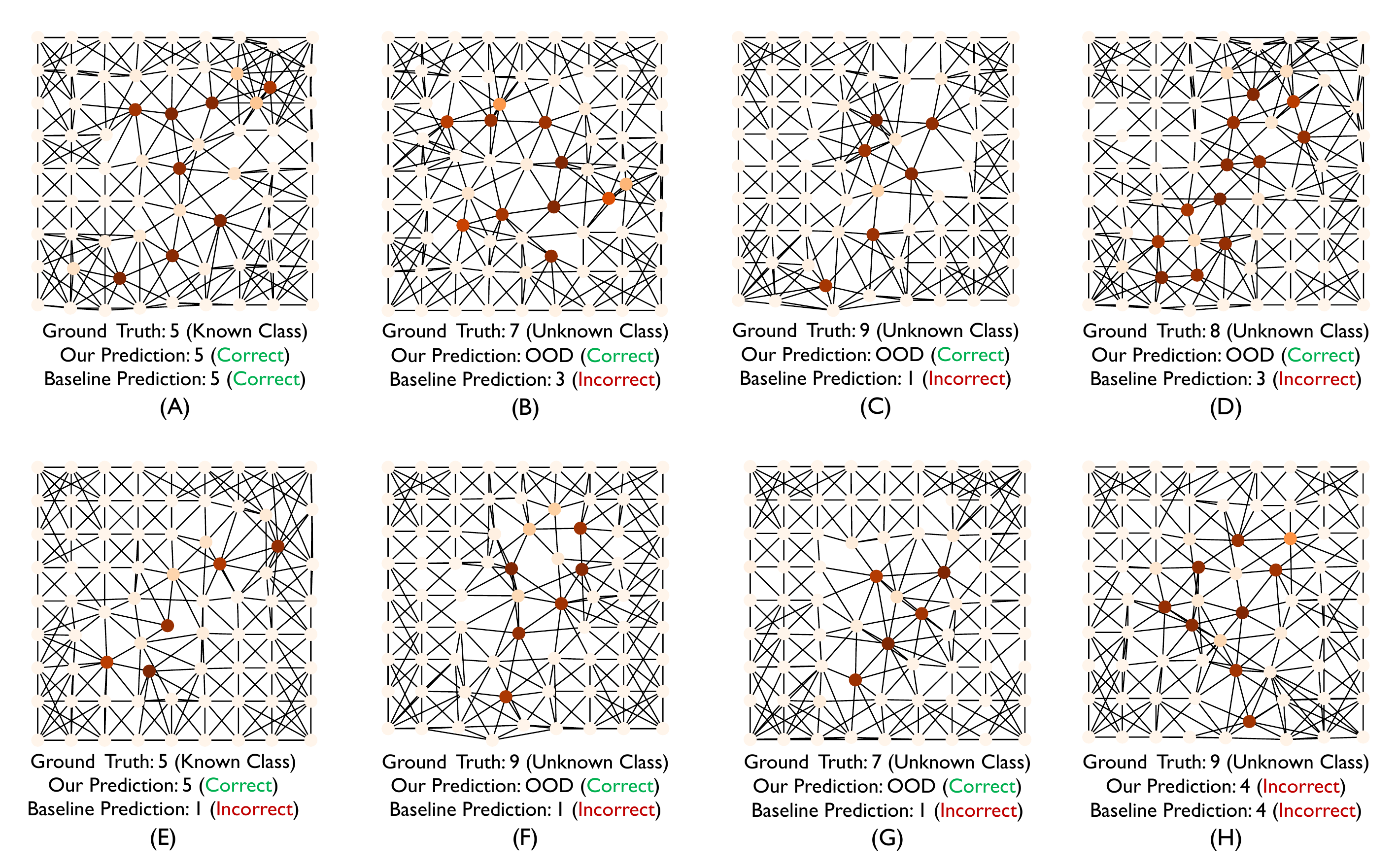}
    \caption{Visualization of classification results. We visualize several graphs in the MNIST dataset with their corresponding ground truth classes, our prediction results and the prediction results of the baseline model using GIN and TopK pooling. The results show that in many cases, the proposed \method{} correctly detects OOD samples.}
    \label{fig:case}
\end{figure*}
\subsection{Parameter Sensitivity (RQ~3)}
In this subsection, we explore the model's sensitivity to hyperparameters. Specifically, we focus on three hyperparameters: (\romannum{1}) the number of clusters (prototypes of unknown classes $R$) in Prototype-aware Semi-supervised Learning, (\romannum{2}) the number of presumed OOD samples, and (\romannum{3}) the temperature in Eq.~\ref{eq:contrast}. Note that in the previous experiments, we use the ground truth number of OOD samples and the goal of the experiment (\romannum{2}) is to show that our \method{} is not sensitive to the presumed number of OOD samples. For mathematical convenience, we introduce a coefficient $\alpha$ called OOD sample multiplier, and instead of choosing top $K$ samples, we select top $\alpha K$ samples. The experimental results are shown in Fig.~\ref{fig:sensitivity}.

\R{As can be seen from the results, although the performance fluctuates around $R\in [4,5]$ in some cases, fixing the number of clusters $R$ to 3 achieves the best performance.} \R{In particular, too few clusters ($R<3$) would weaken the benefit of prototype learning on OOD samples while too many clusters ($R>3$) would make the model overfitting after saturation.} Moreover, while the actual number of unknown classes varies (\emph{i.e.} 20 unknown classes for the COIL-DEL dataset and 5 unknown classes for the Letter-High dataset), assuming a moderate number of unknown prototypes (clusters) is generally beneficial for the model. One explanation is that the model uses K-means algorithm to initialize the unknown prototypes, which often yield balanced results with three clusters. Another possible reason for this phenomenon relates to the nature of OOD discovery. The model has little knowledge of out-of-distribution samples (since they are all unlabeled) and can only resort to the internal structures and attributes of the graphs to calculate the feature of each sample. This leads to coarse representations of OOD samples, and clustering them into too many prototypes becomes very challenging.

As for the number of OOD samples, the experiments demonstrate that an accurate estimation is not required to achieve competitive results. As can be seen from the middle column of Fig.~\ref{fig:sensitivity}, perturbing the presumed number of OOD samples in the range of 80\% to 140\% has little influence on the results. An interesting finding from the experiments is that the overestimation of the number has less influence compared to underestimation. One possible reason for this is that in the semi-supervised setting, there could be some samples belonging to the known classes but very different from labeled data. For example, there could be several sub-classes for a known class, but one sub-class does not appear in the labeled data. To some extent, this sub-class plays the role of an unknown class, and it is beneficial for the model to identify this fact.

For the temperature parameter, we can observe that our \method{} achieves the best accuracy when the temperature is set to 0.1, providing the correct ``softness'' for the softmax function in Eq.~\ref{eq:contrast}. An interesting phenomenon is that in the low labeling ratio case, the model is more sensitive to changes in the temperature parameter. The possible reason is that with less labels as supervision, the model relies relatively more on semi-supervised learning, and the influences of hyperparameters in Prototype-aware Semi-supervised Learning get amplified.

\subsection{Visualization (RQ~4)}
\subsubsection{Visualization of Classification Results}
In this subsection, we visualize eight graphs and our predictions in comparison with the prediction of the baseline model. Specifically, the experiments are conducted on the MNIST dataset, and we use GIN convolution~\cite{GIN} with TopK pooling~\cite{topk-pooling} as the baseline. The result is shown in Fig.~\ref{fig:case}, and from the results, we have several observations:
\begin{itemize}
    \item The task is generally more challenging than hand-written digits recognition in images, and GNN can perform relatively well. For example, in case (A), although the structure of the digit `5' is not very clear, both baseline model and the proposed \method{} classify this graph correctly.
    \item The baseline model is weak in detecting OOD samples, while the proposed \method{} is better at finding OOD samples. For example, in case (B), (C), (D), (F) and (G), the graphs belong to the unknown classes and the models are not provided with corresponding labels during training. In these more challenging cases, the proposed \method{} detects OOD samples correctly, whereas the baseline model yields seemingly reasonable but incorrect predictions.
    \item The proposed model is also better at classifying known classes. For example, in case (E), the graph belongs to the known class and while the baseline model fails to classify it correctly, our \method{} yields the right answer. This suggests that the proposed Subgraph-based OOD Detection and Prototype-aware Semi-supervised Learning not only help with finding out-of-distribution samples but also improve the representations of in-distribution samples.
    \item There are some very hard cases where both the proposed \method{} and the baseline model fail. For example, in case (H), the ground truth is `9', but the models do not see graphs with label `9' during training. To make things worse, it resembles the digit `4', which is often seen during training with labels. To some extent, it is reasonable for the model to make such mistakes.
\end{itemize}

\subsubsection{Visualization of Learned Representations}
\begin{figure}[t]
    \centering
    \includegraphics[width=0.49\textwidth,keepaspectratio=true]{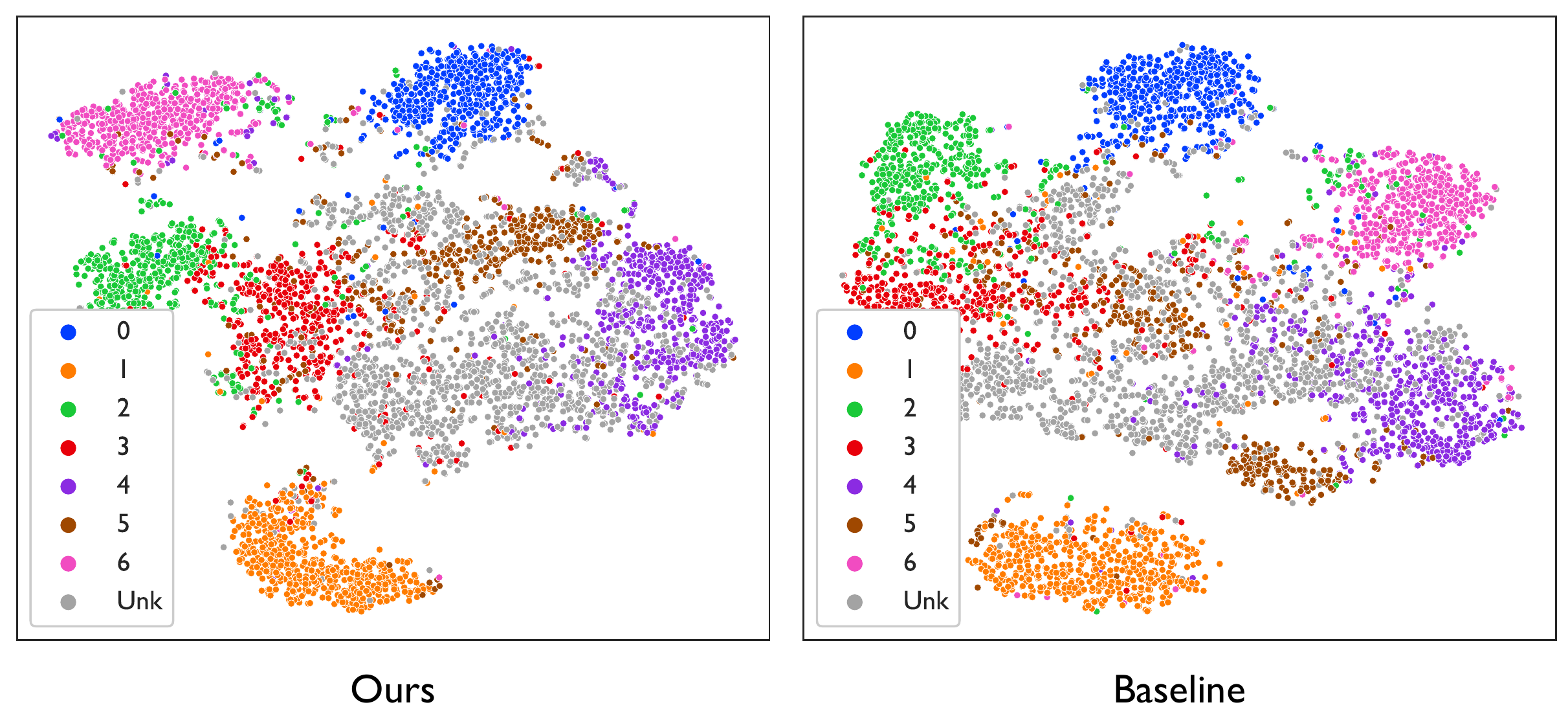}
    \caption{Visualization of learned graph features using t-SNE. The experiments are performed on the MNIST dataset with high labeling ratio, comparing the learned representations of our model (left) and a baseline model using GIN convolution and TopK pooling (right). Our model achieves the overall accuracy of 73.04\%, while the baseline model reaches 62.55\% overall accuracy. The results show that \method{} yields better representations.}
    \label{fig:tsne}
\end{figure}
We use t-SNE~\cite{t-SNE} to visualize the results of learned representations, which is shown in Fig.~\ref{fig:tsne}. More specifically, the experiments are conducted in the high labeling ratio case of the MNIST dataset, and we compare the results of \method{} in comparison with a GNN baseline that uses GIN convolution~\cite{GIN} and TopK pooling~\cite{topk-pooling}. As can be seen from the results, our learned representations are more condensed. For example, for the class of digit 3 (red dots), our model yields more condensed features that are less confused with class Unk (unknown classes, gray dots). Another example is the class of digit 5 (brown dots), which is cut into two clusters (one cluster in the middle of the graph close to the red dots, the other lower in the graph between the purple dots and the orange dots) in the results of the baseline model. In comparison, our learned representations are better in that most brown dots are clustered into one group. 

For the unknown classes, we find it challenging to distinguish their features with other learned representations of known classes clearly. However, our results are better than the baseline's. As can be seen in Fig.~\ref{fig:tsne}, the proposed \method{} not only provides a more condensed representation distribution of OOD samples, but also sets clearer boundaries. For example, our model better separates the OOD samples with the class of digit 2 (green dots). 

We attribute the more condensed representations and clearer boundaries among classes to the Subgraph-based OOD detection and the following Prototype-aware Graph Semi-supervised Learning. Compared to the baseline model that uses only cross entropy loss, our method can better capture the internal structure of the graphs that tend to distinguish themselves from other samples.

%% file: 6_conclusion.tex
\section{Conclusion}\label{sec:conclusion}
This research studies the topic of semi-supervised universal graph classification, which attempts not only to detect graph samples that do not correspond to known classes but also to classify the remaining samples into their respective classes. From a subgraph prospective, we offer a novel approach dubbed \method{} that overcomes both class shifts and label scarcity in this problem. On the one hand, to achieve reliable OOD sample detection, \method{} samples several subgraphs for each sample and then measures both prediction confidence and individual output uncertainty comprehensively. On the other hand, \method{} builds graph prototype representations and then use the posterior prototype assignments inferred from one subgraph view to monitor the semantics of unlabeled input from another view. Extensive experiments on four benchmark graph classification datasets demonstrates the efficacy of our \method{}. In future work, we will apply our \method{} to more realistic graph classification scenarios, including domain adaptation and domain generalization.